\documentclass{article}
\usepackage{PRIMEarxiv}
\usepackage[utf8]{inputenc} 
\usepackage[T1]{fontenc}    
\usepackage{hyperref}       
\usepackage{url}            
\usepackage{booktabs}       
\usepackage{amsfonts}       
\usepackage{nicefrac}       
\usepackage{microtype}      
\usepackage{lipsum}
\usepackage{adjustbox}
\usepackage{siunitx}
\usepackage{caption}
\usepackage{graphicx}       
\graphicspath{{media/}}     
\usepackage{wrapfig}
\usepackage{xcolor}
\usepackage{longtable}
\title{Large Language Models estimate fine-grained human color-concept associations}

\author{
  Kushin Mukherjee \\
  Psychology and Wisconsin Institute for Discovery\\
  University of Wisconsin-Madison \\
  Madison, WI. USA\\
  \texttt{kmukherjee2@wisc.edu} \\
  \\
  \And
  Timothy T. Rogers\\
  Psychology and Wisconsin Institute for Discovery\\
  University of Wisconsin-Madison \\
  Madison, WI. USA\\
  \texttt{ttrogers@wisc.edu} \\
  \AND
  Karen B. Schloss \\
  Psychology and Wisconsin Institute for Discovery\\
  University of Wisconsin-Madison \\
  Madison, WI. USA\\
  \texttt{kschloss@wisc.edu}
}

\begin{document}
\maketitle

\begin{abstract}

People have measurable associations between concepts and colors, even for concepts that are highly abstract or those that do not denote items with diagnostic colors.
Concepts elicit a distribution of association strengths across perceptual color space, and such distributions are known to influence aspects of visual cognition ranging from object recognition to interpretation of information visualizations. While prior work has hypothesized that color-concept associations may be learned from the cross-modal statistical structure of experience, it has been unclear whether natural environments possess such structure or, if so, whether learning systems are capable of discovering and exploiting it without strong prior constraints. 
 We addressed these questions by investigating the ability of GPT-4, a multimodal large language model, to estimate human-like color-concept associations without any additional training. Starting with human color-concept association ratings for 71 color set spanning perceptual color space (\texttt{UW-71}) and concepts that varied in abstractness, we assessed how well association ratings generated by GPT-4 could predict human ratings. GPT-4 ratings generated from purely natural language prompts were  correlated with human ratings, with performance comparable to state-of-the-art methods for automatically estimating color-concept associations from images. Variability in GPT-4's performance across concepts could be explained by specificity (peakiness) of the concept's color-concept association distribution. This study suggests that high-order covariances between language and perception, as expressed in the natural environment of the internet, contain sufficient information to support learning of human-like color-concept associations, and provides an existence proof that a learning system can encode such associations without initial constraints or biases to do so. The work further demonstrates that GPT-4 can be used to efficiently estimate distributions of color associations for a broad range of concepts, potentially serving as a critical tool for designing effective and intuitive information visualizations.
\end{abstract}

\section*{Introduction}

 It is no surprise that people connect the names of familiar objects with their characteristic colors---for instance, strongly associating `carrot' with shades of orange. Recent work suggests, however, that such associations are not limited to characteristic colors of concrete concepts; instead, people can reliably judge association strength between essentially any concept and any color, regardless of the concept's concreteness or the extent to which it possesses characteristic colors (Figure \ref{fig:example}). Thus, any given concept can be viewed as eliciting a {\em distribution} of association strengths across all of color space--a phenomenon that in turn constrains how color meanings are interpreted in various forms of visual communication \cite{schloss2021, mukherjee2021context, schloss2024color}. Prior research has shown that color-concept association distributions for novel object concepts can be learned from the covariance structure of the environment \cite{schoenlein2022colour}, but how might these associations be acquired for concepts that do not clearly encompass concrete, perceptible items in the world? 
One possibility is that such associations are latent in the statistical structure of experience across language and perceptual experience. Just like for concrete concepts, patterns of co-occurrence between words and phrases, observed objects, and the events or scenarios in which these arise may lead a learner to connect perceived colors to even quite abstract and non-visually observable concepts. While this hypothesis has been advanced in prior work \cite{schlossPICS, soriano2009emotion}, it has been unclear whether such correlational structure exists in real-world language and experience, and if it does, whether any learning system is capable (without prior constraints) of detecting and exploiting it in human-like ways.

This paper uses contemporary advances in artificial intelligence (AI) to answer these questions. We assessed a state-of-the-art large language model (LLM), GPT-4 \cite{openai-gpt4}, in its ability to generate human-like patterns of color-concept associations.
Such an approach may seem counter-intuitive because LLMs are primarily trained on written language; thus it may seem that they could, at best, generate associations between concepts and colors that are nameable, rather than to all possible perceived colors. Yet recent work has shown that the web-scale datasets used to train contemporary LLMs implicitly encompass an analog of cross-modal learning \cite{openai2023gpt4, bubeck2023sparks}.
Specifically, text-based languages for rendering visual images (e.g. support-vector graphics) include standards for referring to highly specific colors, such as hexadecimal-based color descriptors. Such `color tags' are embedded within code used to render images on web pages, which in turn is embedded within the natural language describing accompanying text on the page. 
On this basis, LLMs can complete a remarkable range of tasks that depend on knowledge and/or perception of color---for instance, generating vector graphics of animals and scenes using text descriptions as input \cite{bubeck2023sparks} or providing judgements about perceptual similarity \cite{marjieh2023language}.
These findings suggest that LLMs trained on web-scale corpora can provide a vehicle for assessing whether color-concept associations can be learned, even for quite abstract concepts, from complex patterns of co-variation between language and textual signifiers of color. 

For this reason we conducted a comparative study of color-concept association rating in humans and in GPT-4, evaluating the model on three criteria. 
First, it should be capable of estimating, not just the most strongly-associated color for a given concept, but distributions of association-strengths across a set of colors spanning color space. 
Second, for any given concept, the model-generated association strengths across colors should correlate with the mean association strengths observed in human judgments. 
Third, the model should be able to estimate these association distributions for any concept, regardless of its level of concreteness or the degree to which the concept is strongly associated with a small number of highly diagnostic colors.

From a theory standpoint, meeting these criteria would provide evidence that (1) the natural environment of the internet contains sufficient statistical structure to support color-concept association, and (2) there exists a learning system capable of discovering and exploiting such structure without strong prior biases guiding it to do so. 
From a practical standpoint, meeting the criteria would suggest that GPT-4 and related models can provide a resource-efficient means for estimating color-concept associations, which in turn provides a basis for optimizing color use in visual communication \cite{lin2013, heer2012color,schloss2020semantic,mukherjee2021context}. 
In the rest of this introduction we briefly review previous efforts to automatically estimate color-concept associations to provide context for the current work. We then report three experiments comparing human and LLM-generated color-concept associations, and in the general discussion, consider the implications of the results both for a cognitive theory of color semantics and for practical application to designing effective information visualizations \cite{schloss2024color}.

\subsection*{Previous approaches to automatically estimating color-concept associations}  

Methods have been developed to automatically estimate color-concept associations that have been used for designing information visualizations with easily interpretable colors \cite{lin2013, rathore2019estimating, bartram2017}. These approaches have leveraged  visual knowledge latent in large-scale language databases, image databases, or a combination of both.   
Techniques based in language exploited co-occurrence frequency between single concept words and color information in large corpora \cite{havasi2010,lindner2012large, bartram2017, rathore2019estimating, lin2013}. For example, Havasi et al. \cite{havasi2010} used a semantic network (ConceptNet \cite{speer2017conceptnet,liu2004conceptnet}) to find associated color terms for concept words. Then, they used a color-naming database (XKCD color survey dataset \cite{XKCDBlogColorSurvey}) to map the color terms to a perceptual color space and determined the `best' associated color as the centroid of the associated colors. 

Subsequent approaches used language corpora to map concepts to basic color terms  and then linked those color terms to colored patches using color information in labeled image databases \cite{setlur2016} with a method established by \cite{lin2013}.
Other approaches have also combined information  latent in natural images and language corpora using topic modeling approaches \cite{jahanian2017colors} and using color distributions associated with words to establish the semantic relatedness between concepts \cite{guilbeault2020color, mukherjee2022color}.
When estimating color-concept associations from images, it is necessary to specify which pixel input should ``count'' towards an association between a given color and concept. For example, when evaluating images of blueberries to estimate associations colors and the concept \textit{blueberry}, should an algorithm only consider the exact pixel colors in the image, or should there be some ``tolerance'' around the pixel colors? Rathore et al. \cite{rathore2019estimating} found that the method for best estimating associations that matched human judgments used tolerance that scaled in size along perceptual dimensions of color and spread to all colors that shared the same color category as the pixel input--category extrapolation (e.g., a observing a grayish blue pixel in an image of blueberries counted for all colors categorized as \textit{blue}). 
Bypassing the issue of pixel tolerance, Hu et al. \cite{hu2023} finetuned a pretrained neural network model to predict pixel colors of grayscale images (colorization) and used it to estimate color-concept associations.
Their method was effective for capturing the overall scale of association values, but was overall less effective than previous approaches for capturing the \textit{relative pattern} of associations over color space for a given concept.

Methods to automatically estimate color concept associations have advanced over the past two decades, but several limitations preclude using them to estimate human-like color-concept association distributions for \textit{any} queried concept. 
First, language based approaches returned the top associate or optimal color, but did not provide association distributions over color space. 
This is critically important when designing visualizations because often a concept's most associated color is not the most interpretable color depending on the context \cite{schloss2018color, schloss2020semantic, mukherjee2021context}.
Image-based approaches addressed this limitation by providing distributions of associations, but they pose a different challenge--- accessing the images automatically can be difficult due to the varied and ever-changing interfaces and levels of access to image databases like Google Images. 
Second, implementing specialized computer vision algorithms often takes time and domain-specific knowledge. 
What is needed is a model that jointly represent associations between all colors and all concepts in a common representational space, which can quickly return color-concept association distributions for any concept. 
We propose that Large Language Models have the potential to serve this purpose.

\subsection*{Using language models to study human perceptual knowledge}  
Evidence suggests that LLMs are highly effective at tasks that require natural language, such as writing plausible college-level essays \cite{elkins2020can}, scoring highly on various professional licensing exams \cite{choi2023lawyering, kung2023performance, newton2023chatgpt}, and generating functional code in many programming languages \cite{openai-gpt4}. 
LLMs such as the BERT family of models \cite{devlin2018bert,liu2019roberta,yang2019xlnet, clark2020electra}, Llama \cite{touvron2023llama}, Flan-T5 \cite{chung2022scaling}, and the OpenAI GPT models \cite{brown2020language}, to name a few, have also been successful at aligning with human behavioral results across several psychological domains including language \cite{piantadosi2022meaning, tuckute2023driving, futrell2019neural,lampinen2022can,kumar2022reconstructing} and perception \cite{marjieh2023language,conwell2023language, mirchandani2023large}.
Even simpler neural networks can learn the structure of \textit{perceptual} spaces by training on large image and language corpora. 
For example, color space coordinates can be decoded from the intermediate representations of autoregressive transformer models \cite{vaswani2017attention} when presented with color terms in natural language \cite{abdou2021can, patel2021mapping}. 

GPT-4 \cite{openai-gpt4} has been particularly successful at capturing human similarity judgements for perceptual features such as color, sound, and touch \cite{marjieh2023language}.
This is possible because LLMs operate over `tokens', which usually stand in for words in natural language, but can also be used to represent perceptual inputs and outputs as long as those perceptual inputs/outputs can be represented in text form.  
The performance of these models on perceptual tasks can be further improved via `task prefixing' \cite{raffel2020exploring}, which involves appending a sequence of tokens to an input prompt that specifies the exact nature of the task that needs to be performed by the LLM in the form of instructions in natural language.
Performance is further improved by `in context learning', which involves providing an example of the task that needs to be performed in the input prompt \cite{brown2020language, mirchandani2023large}.

Thus, it may be possible to use GPT-4 to estimate human-like color-concept associations using methods similar to prompting color-concept associations from humans by (1) specifying colors using language (e.g., hexadecimal codes) and (2) using task pre-fixing to ask the model to provide a numeric `rating' of color-concept associations. 
If this method is effective, it will greatly reduce the effort and difficulty in curating color-concept association ratings while providing support for the notion that color-concept associations can be learned by a sufficiently powerful learning model from the statistics of concept and color information co-occurrences.
Specifically, if GPT-4 through its pretraining is already effective, the difficult hurdle of fine-tuning models to learn color-concept associations between specific concepts and colors can be sidestepped.
Additionally, the GPT-4 family of models can be accessed via easy-to-use APIs, which would provide an efficient means of using color-concept associations for designing effective visualizations and for conducting studies on the role of color in human cognition.

\section*{EXPERIMENT 1}

Experiment 1 assessed how well human color-concept associations could be predicted by color-concept associations generated by GPT-4. The concepts spanned a wide variety of semantic domains, including concrete and abstract concepts (Table \ref{tab:concepts}). The colors were the \texttt{UW-71} colors, which systematically sampled colors across CIELAB space (Figure \ref{fig:uw71})\cite{mukherjee2021context}. The human data were previously  collected in Mukherjee et al. \cite{mukherjee2022color}, in which participants judged associations between concepts and patches of color. We collected new GPT-4 judgments using a similar task given to humans, but GPT-4 judged associations between concepts and colors prompted as hexadecimal codes rather than patches of color.

\subsection*{Methods}

\begin{wrapfigure}{r}{0.5\textwidth}
 \includegraphics[width=0.5\textwidth]{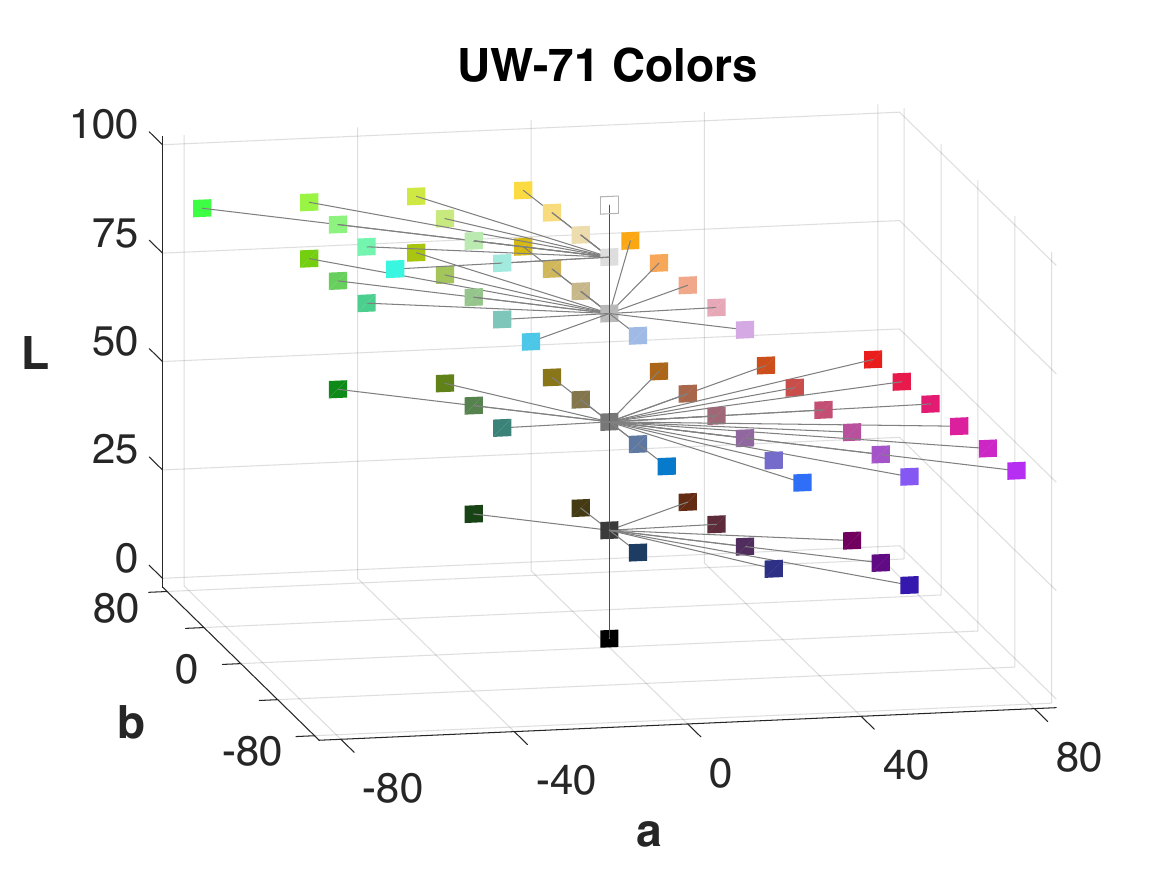}
 \caption{Colors in the UW--71 color library plotted in CIELAB space (figure adapted from Mukherjee et al. \cite{mukherjee2021context}.}
 \label{fig:uw71}
 \vspace{-10mm}
\end{wrapfigure}

\begin{figure}
    \centering
    \includegraphics[width=.9\textwidth]{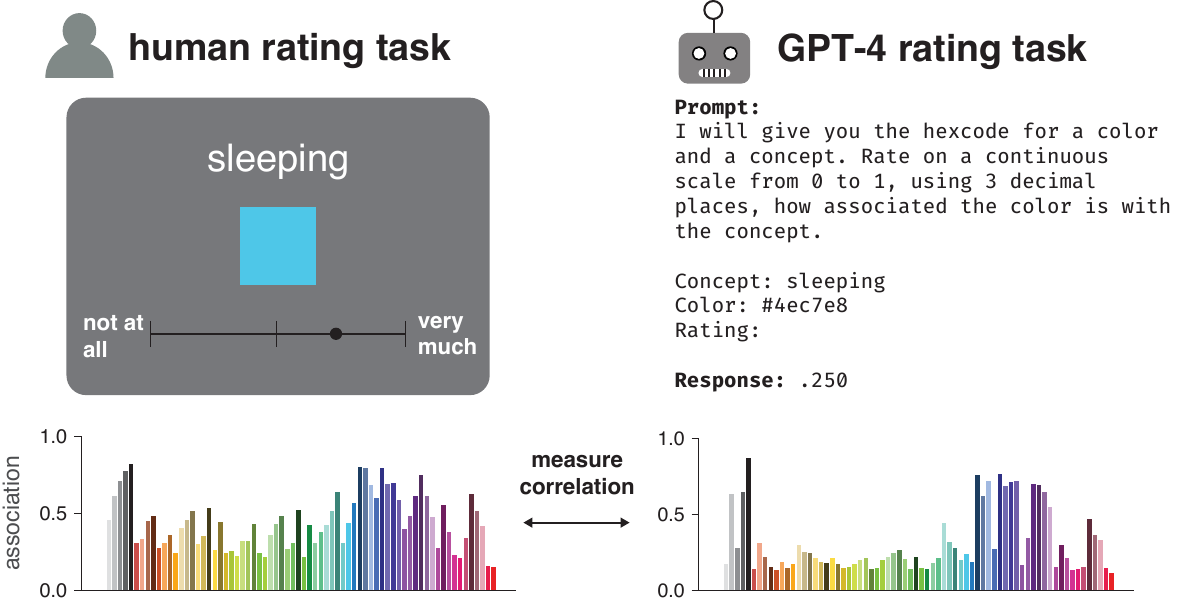}
    \caption{Example trial from the human color-concept association rating task (left) and GPT-4 rating task for the same color-concept pair (right). Bar graphs correspond to average human and GPT-4 color concept association ratings for the concept `bird' over the entire \texttt{UW-71} color library.}
    \label{fig:methods}
     \vspace{-5mm}
\end{figure}

We assessed color-concept associations for the \texttt{UW-71} colors and 70 concepts. The \texttt{UW-71} colors \cite{mukherjee2021context} include 58 colors sampled uniformly in CIELAB space (color distance $\Delta E$=25) \cite{rathore2019estimating}, plus 13 additional colors sampled at higher lightness (L* = 81) to include more ``typical'' yellows excluded from the grid sampling \cite{mukherjee2021context} (Figure \ref{fig:uw71}).
The concepts were organized into 14 categories, with five concepts per category, chosen to span the range from highly concrete (e.g., banana, celery) to highly abstract (e.g., love, happy) (Table \ref{tab:concepts}). Below, we describe the procedure for collecting color-concept associations from humans (previously reported in \cite{mukherjee2022color}), followed by our new procedure for estimating color-concept associations from GPT-4. 

 \subsection*{Data on human color-concept associations.}
Human color-concept association data were collected online by asking participants to rate the association strength between single color patches and single concepts on a sliding scale from `not at all' to `very much' \cite{mukherjee2022color, mukherjee2021context}.
 Before beginning the rating trials, the participants completed an `anchoring task' so they knew what associating `not at all' and `very much' meant in the context of these colors \cite{palmer2013visual}. 
 During anchoring, participants were shown a screen with the concepts they would be asked to rate and all 71 colors. They were instructed to think of which color they most and least associated with each concept. Those colors were to correspond to the extremes of the rating scale.
 
On each rating trial, participants rated how much they associated the given concept with the given color patch by moving the slider marker to the desired location on the response scale and clicking a “continue” button to record their response (Figure \ref{fig:methods}). 
 The display for each trial included a colored patch centered on the screen, with the concept word displayed above.
 The background was a dark gray (CIE Illuminant D65, x = .3127, y = .3290, Y = 10 cd/m2), which enabled perceiving the colors of all lightness levels in the \texttt{UW-71} colors against the background.
Participants judged the colors online on their own devices so the color coordinates were estimates of ``true'' CIELAB color coordinates (see Table \ref{tab:UW_71_colors}). This approach is common among studies of color in information visualization, for which findings need to be robust to natural variations in displays characteristics \cite{stone2014, szafir2018, gramazio2017, mukherjee2021context, schoenlein2022unifying}.

 The color-concept association ratings were collected from a total of 720 participants.
Data for four of the 14 concept categories (fruits2, vegetables, properties, and activities) were collected as a part of Experiment 2 in Mukherjee et al. (2022) \cite{mukherjee2021context} and data for the remaining concepts were collected in Mukherjee et al. (2022) \cite{mukherjee2022color}.
Refer to the two references above for details regarding participant demographics and methodological details regarding data collection.
Participants in both experiments provided informed consent in accordance with the University of Wisconsin-Madison IRB.

{

\begin{table}
\captionsetup{skip=10pt} 
    \centering
    \begin{tabular}{l|l|l}
    \toprule
        \textbf{Category} & \textbf{Concepts} &\textbf{\textit{n}} \\ 
        \midrule
        Activities & Driving, Eating, Sleeping, Leisure, Working & 52\\
        Animals & Bear, Bird, Lion, Frog, Fish & 45\\
        Automobiles & Airplane, Car, Boat, Truck, Train & 51\\
        Clothes & Dress, Pants, Shirt, Socks, Shoes  & 48\\
        Directions & Above, Below, Beside, Near, Far & 44\\
        Emotions & Angry, Disgust, Fearful, Happy, Sad & 50\\
        Fruits & Blueberry, Lemon, Mango, Strawberry, Watermelon & 46\\
        Fruits2 & Apple, Banana, Cherry, Grape, Peach & 49\\
        Properties & Comfort, Efficiency, Reliability, Safety, Speed,  & 50\\
        Scenes & Beach, Field, Ocean, Sky, Sunset & 45\\
        Times of Day & Dawn, Day, Dusk, Noon, Night & 46\\
        Values & Evil, Greed,Justice, Love,Peace & 50\\
        Vegetables & Carrot, Celery, Corn, Eggplant, Mushroom & 52\\
        Weather & Blizzard, Drought, Hurricane, Lighting, Sandstorm  & 52\\
        \bottomrule
        
    \end{tabular}
    
    \caption{Categories of concepts tested in this study (five concepts per category), and corresponding number of participants (\textit{n}) for each category.}
    \label{tab:concepts}
     \vspace{-5mm}
\end{table}
}

\textbf{GPT-4 color-concept association ratings} 

To obtain estimates of color-concept associations from GPT-4, we adapted the color-concept association rating procedure conducted with human participants. We began by assigning a `system' level prompt to GPT-4 conditioning the model to be \textit{``an expert on color-concept associations''}. 
We then provided two sentences describing the task - \textit{``I will give you the hexcode for a color and a concept word. Rate on a continuous scale from 0 to 1, using 3 decimal places, how associated the color is with the concept.''}. No information was provided about a background color.
Next, we added a sentence indicating the start of a rating trial \textit{``Let's do the rating task —''}, followed by 3 lines, one indicating the concept to be rated, one indicating the hexcode for the \texttt{UW-71} color to be rated and blank field for the rating with text asking the model to \textit{only} answer with a number.
Here is an example of what this procedure  was like for the concept `apple' and the color hexcode `\#FFFFFF':

\begin{verbatim}
Concept: `apple'
Color: #FFFFFF 
Answer with only the number:
\end{verbatim}

Each rating was obtained using this prompt structure, with only the concept and color being swapped depending on the pair being rated on that trial. The temperature was set to 0 to produce deterministic outputs (i.e., multiple prompts for the same condition would result in the same output). Each color-concept pair was rated once (71 colors $\times$ 70 concepts  = 4,970 total ratings).

\subsection*{Results and Discussion}
\begin{figure}[htb!]
    \centering
    \includegraphics[width=1\textwidth]{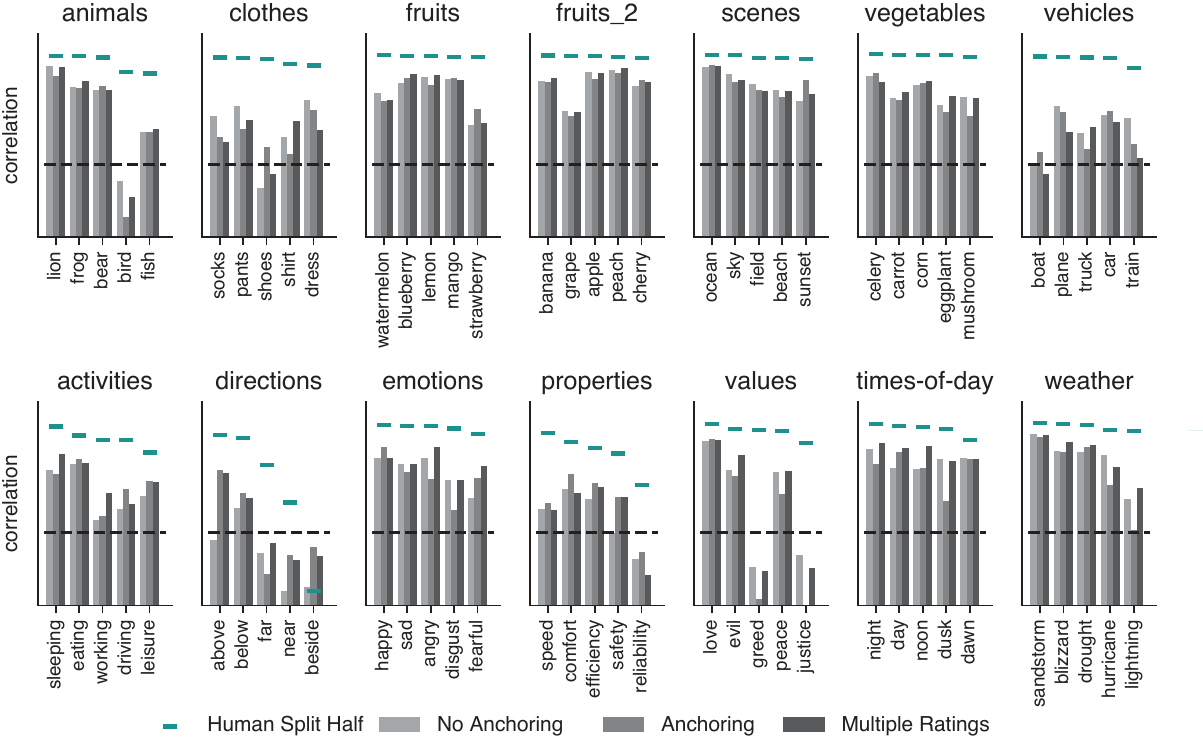}
    \caption{Correlations between average human color-concept association ratings and predicted ratings from GPT-4 from experiments 1 (light grey bars), 2 (medium gray bars), and 3 (dark grey bars) for each of our 70 concepts. Teal lines above each concept correspond to the average human split-half reliability for that concept.}
    \label{fig:cors}
\end{figure}

To assess the efficacy of GPT-4 for estimating human color-concept association ratings, we first set a human performance ``benchmark'' for comparison using split-half correlation. 
For each concept, we randomly divided the human participants into two groups, calculated the mean association for each color within the two groups, and computed the correlation over association ratings between the two groups for that concept. 
After applying the Spearman-Brown correction to this correlation value it constitutes the split-half correlation for the concept for a given split of the data.
We repeated this procedure 50 times, randomly splitting the data on each iteration, and averaged the split-half correlations to obtain a mean split-half correlation for each concept. This correlation provided a ceiling against which model-generated ratings could be assessed.
Figure \ref{fig:cors} shows the mean split-half correlations as blue horizontal lines above each concept. 

Next, for each concept, we computed the correlation between GPT-4 estimates and mean human association ratings across colors. These results are shown in Figure \ref{fig:cors} as vertical light gray bars labeled ``no anchoring.''  
We refer to this task as the ``no anchoring'' version of the GPT-4 rating task since we did not prompt for the anchoring trials that humans performed before rating associations.
Overall across all concepts, there was a moderately strong correlation between GPT-4 and human ratings (mean $r$(69)= .67), although there was variability in correlations across concepts (max $r$ = .93, min $r$ = .08). The horizontal gray bars indicate the $r$ value above which a correlation was considered statistically significant after correcting for conducting 70 correlations corresponding to the 70 concepts (Bonferroni adjusted critical $\alpha = .0007$). The correlations exceed this $r$ value for 59 out of 70 concepts. For some concepts, GPT-4 correlations come close to the split-half correlation baseline (e.g., lion), but for other concepts GPT-4 fell short (e.g., boat). A paired samples t-test over concepts indicated that GPT-4 correlations (mean $r$(69) = .67) were significantly lower than the split-half correlations (mean $r$(69) = .93), $t$(69) = 13.72, $p$<.001). 

\begin{figure}
    \centering
    \includegraphics[width=1\textwidth]{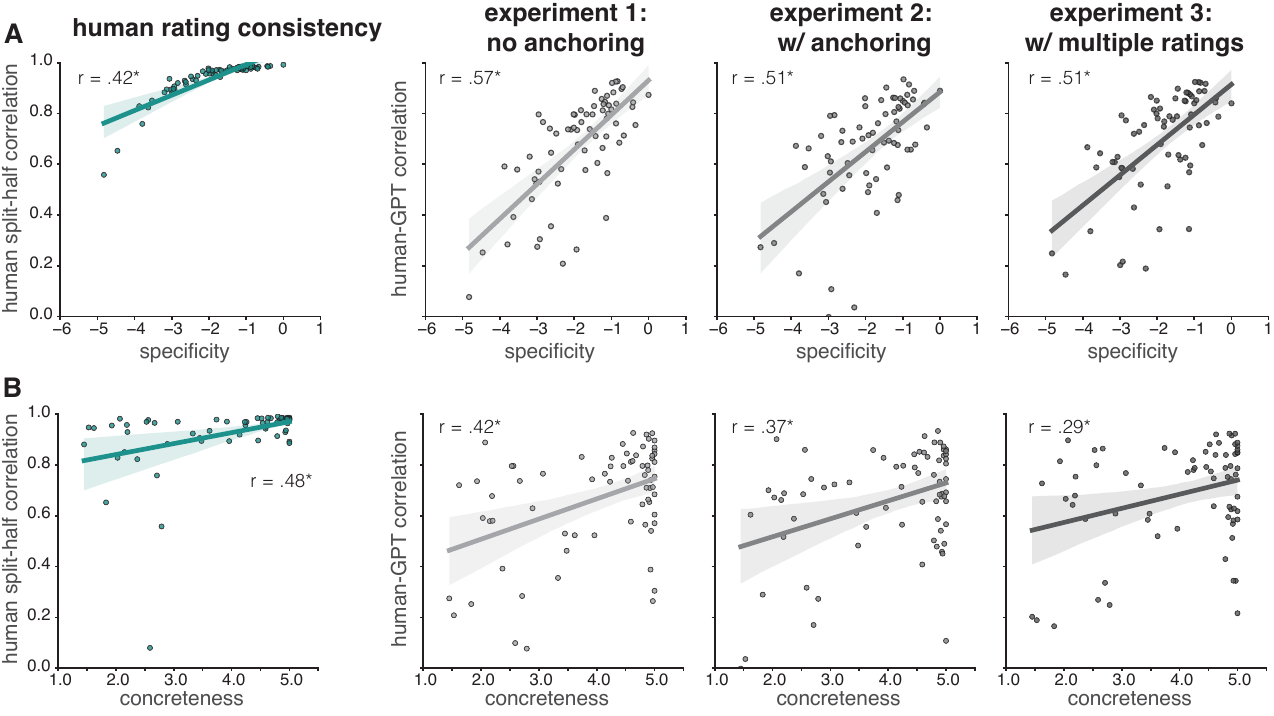}
    \caption{(A) Relationship between the specificity of concepts' color-concept associations and GPT-4's ability to accurately predict associations. Each point represents a different concept. (B) Relationship between the concepts' concreteness and GPT-4's ability to accurately predict associations. The first column in (A) and (B) reflect human split-half correlations for color-concept associations as a function of specificity (top) and concreteness (bottom).}
    \label{fig:spec_v_cor}
\end{figure}

These results are comparable to other state-of-the-art methods for estimating color-concept associations that require at least \textit{some} supervised training or data curation\cite{rathore2019estimating,hu2023}. Table  \ref{tab:other_datasets} shows our results in comparison to results  from Rathore et al. \cite{rathore2019estimating} and Hu et al. \cite{hu2023} where there overlapping concepts across studies. 
First, we compared the average quality of the estimated associations, operationalized using correlation, between our current method and that of Rathore et al. for a set of five fruit concepts common to both studies -- blueberry, lemon, mango, strawberry, and watermelon. The mean correlation between the true and estimated associations using Rathore et al.'s method for the 5 fruits was $r$(57) = .79 (max $r$(57) = .92; min $r$(57) = .65)\footnote{This study used the \texttt{UW-58} color set, which is a subset of the current \texttt{UW-71} color set}, which is comparable to our results using GPT-4 (mean $r$(70) = .80; max $r$(70) = .90; min $r$(70) = .61).
We conducted a similar comparison with the results from Hu et al. \cite{hu2023} for a set of 14 concepts (fruits and vegetables) common to both our studies.
The correlations for our method for these concepts were on average higher (mean $r$(70) = .80; max $r$(70) = .90; min $r$(70) = .61) than the correlations from Hu et al.'s method based on colorization neural networks (mean $r$(70) = .71; max $r$(70) = .88; min $r$(70) = .12). Here, we focus on correlation with human ratings as a the key metric for assessing the quality of estimated associations due to the importance of relative patterns of associations across colors, but we note that Hu et al. investigated alternative metrics for evaluating their estimates (e.g.,  earth mover's distance (EMD) between distributions).
\begin{figure}[htb!]
    \centering
    \includegraphics{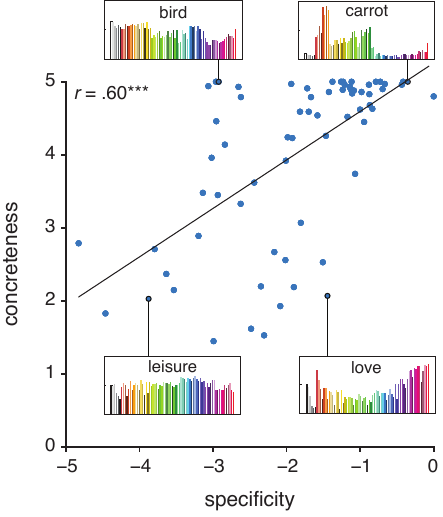}
    \caption{The relationship between specificity and concreteness for the set of 70 concepts.
    The color-concept association distributions for each of the concepts bird, carrot, leisure, and love are plotted on a scale from 0 to 1.}
    \label{fig:example}
\end{figure}
To investigate why there is a large range of correlations over concepts, we considered that a concept's distribution specificity may play a role. That is, concepts that have stronger, specific associations may be easier to predict. To test this possibility, we computed the specificity of the color-concept associations for each concept, following the methods in Mukherjee et al. \cite{mukherjee2021context}. First, we computed the inverse of the Shannon entropy of its associations over the full set of \texttt{UW-71} colors. Then, we normalized the entropy values to be in the 0-1 range across all concepts and computed the log of its additive inverse to linearize the data. 
The specificity of each concept correlated significantly with human split-half correlations (Figure \ref{fig:spec_v_cor}A leftmost panel; $r$(69) = .42, $p$ < .001), indicating that human ratings were more reliable across participants for concepts with more specific distributions of color associations. 
The same trend held for the GPT-4 association ratings from this experiment in a more pronounced manner (Figure \ref{fig:spec_v_cor}A second panel). As specificity increased, so did human-GPT correlations across the \texttt{UW-71} colors ($r$(69) = .57, $p$ < .001).
Thus GPT-4 estimates were less aligned for the same concepts for which human inter-subject agreement was lower.

We also considered the possibility that GPT-4 might be worse at estimating associations for concepts that are abstract, partly due to their not possessing any observable colors that would be reflected in the models' training corpora.
To test this hypothesis we used word concreteness ratings collected by Brysbaert et al. \cite{brysbaert2014concreteness} and tested whether a concept's concreteness was predictive of how well GPT-4 could estimate human-like associations. A concept's concreteness was significantly correlated with both human split-half correlations ($r$(69) =.48, $p$ < .001) and human-GPT association correlations ($r$(69) = .42, $p$ < .001) (Figure \ref{fig:spec_v_cor}B). 
Guilbeault et al. \cite{guilbeault2020color} previously showed that for a small set of concrete and abstract concepts abstractness was inversely related to the variability of colors present in images of the concepts from web corpora (specificity).
We found a similar relationship in our data, with a significant positive correlation between concreteness and specificity ($r$(69) = .60, $p$ < .001).
We do note, however that concepts like `bird' and `carrot' can be high in concreteness yet vary in specificity and likewise concepts like `leisure' and `love' are low in concreteness yet vary in specificity (see Figure \ref{fig:example}).

However, when both concreteness and specificity were included in a multiple regression model predicting human-GPT correlations,  specificity was a statistically significant predictor ($t$ = 3.93, $p$ < .001) while concreteness was not ($t$ = 1.43, $p$ = .16).
These results suggest that the effect of concreteness on human-GPT correlations is subsumed by the effect of specificity.

In summary, Experiment 1 suggests that color-concept associations estimated via GPT-4 (using one-shot zero temperature behavior) correlate significantly with human-generated associations for the majority of concepts tested.
The fidelity of such predictions to human ratings depends on the specificity of the color-concept association distributions calculated from human judgments. For some concepts, GPT-4-generated estimates approached human split-half reliability estimates, but for most concepts, human-machine correlations were lower than human-human correlations.
These results suggest that using GPT-4 in a zero-shot manner can be an effective method to estimate human color-concept associations, but efficacy depends on specificity of the concept's association distribution.

{
\begin{table}[htb!]
\captionsetup{skip=10pt} 
    \centering
    \begin{tabular}{l|c|c|c}
    \toprule
        \textbf{Category} & \textbf{Hu et al. \cite{hu2023}} &\textbf{ Rathore et al. \cite{rathore2019estimating}} & \textbf{GPT-4 (current study)}\\ 
        \midrule
        Apple & .69&N.A.& .90\\ 
        Banana & .88&N.A.& .84\\ 
        Blueberry & .84 & .84& .84\\
        Carrot & .82 &N.A.& .75\\
        Celery & .77 &N.A.& .87\\
        Cherry & .62&N.A.& .82\\
        Corn & .81&N.A.& .83\\
        Eggplant & .49&N.A.& .71\\
        Grape & .12&N.A.& .69\\
        Lemon & .92 & .92 & .87\\
        Mango & .92 & .92 &.86\\
        Mushroom & .54 &N.A. &.76\\
        Peach & .86 &N.A.&.90\\
        Strawberry & .66 &.66 & .61\\
        Watermelon & .65 & .65 &.78\\ 
        
        \bottomrule
 
    \end{tabular}

    \caption{Correlations between human color-concept associations and estimated associations using the methods described in Hu et al.\cite{hu2023}, Rathore et al. \cite{rathore2019estimating}, and the current GPT-4 study. N.A.s refer to concepts that were absent in a dataset}
    \label{tab:other_datasets}
\end{table}
}
\section*{EXPERIMENT 2}

Experiment 2 sought to improve GPT-4's performance by adapting the prompt to provide more context for the rating task \cite{chung2022scaling}. Recall that humans initially completed an anchoring task to know what ``not at all'' and ``very much'' meant in the context of the particular colors and concepts judged. 
In Experiment 1, GPT-4 was given no such anchoring task. In Experiment 2, we assessed whether model performance would improve by giving similar guidance in the model prompt.

\subsection*{Methods}
 We evaluated GPT-4 using the same procedure as Experiment 1, but with a modified prompt structure.  
Specifically, each trial began with two additional sentences prior to the rating task as follows (using `apple' as an example):
\begin{verbatim}
The concept is `apple'. 
Before rating, here's the set of all the colors {all_color_hexcodes}. 
Think of which color you associate most with `apple.' That color should get a rating of 1. 
Now think of which color you associated least with `apple.'  
That color should get a rating of 0. Now let's do the rating task.
\end{verbatim}

Here, \texttt{all\_color\_hexcodes} refers to a comma-separated list of all 71 color hexcodes for the \texttt{UW-71} color library.
Like Experiment 1, the temperature parameter was set to 0 and we collected a total of 4,970 ratings.

\subsection*{Results and Discussion}
Figure \ref{fig:cors} shows the GPT-4 correlations in Experiment 2 for each concept as medium gray bars, labeled ``anchoring.'' As in Experiment 1, correlations with human ratings for the majority of concepts were positive and statistically reliable.  The correlations exceeded the critical $r$ value for 63 out of 70 concepts. 

However, as in Experiment 1, the mean correlation across all 70 concepts in Experiment 2 (mean $r$ (70) = .66) was lower than the mean of the human split-half correlations (mean $r$(70) = .93) ($t$(69) = 12.19, $p$<.001). Adding anchoring in Experiment 2 did not significantly improve the GPT-4 correlations (mean $r$(70) = .66) relative to Experiment 1 (mean $r$(70) = .67) ($t$(69) = .60 $p$ = .55). Although the correlations for some concepts in Experiment 2 increased compared to Experiment 1 (see concept `above' in Figure \ref{fig:cors}), others decreased (see concept 'night'). 

Also as was in Experiment 1, both specificity ($r$(69) = .51, $p$ < .001) and concreteness ($r$(69) = .37, $p$ < .05) were significantly correlated with how well GPT-4 was able to capture human-like association ratings, but when both factors were entered into a linear regression model, specificity was a significant predictor ($t$ = 4.48, $p$ < .001) and concreteness was not ($t$ = .03, $p$ = .25).

In summary, Experiment 2 showed GPT-4 performed similarly when prompted with an anchoring than when it was given no anchoring, which suggests that the additional cost and effort for anchoring is does not produce a measurable improvement.

\section*{EXPERIMENT 3}

In Experiments 1 and 2, we set GPT-4 to be deterministic (temperature = 0) so we only prompted it one time for each color-concept pair. However, human color-concept associations are stochastic, which is why we collect color-concept association data over many participants and average their associations to get a stable measure, rather than collecting data from one participant. Thus, we considered the possibility that making GPT-4 more stochastic (raising the temperature) and averaging multiple judgments for each color-concept pair could color produce estimates that better match human color-concept associations. We tested this possibility in Experiment 3.  

\subsection*{Methods}

Experiment 3 followed the same procedure as Experiment 1, but setting the temperature parameter to 1 and rating each color-concept pair 10 times (without anchoring). 
This approach allowed us to simulate the process of acquiring multiple ratings per participant and averaging across participants to acquire mean association ratings per color-concept pair. We collected a total of (70 concepts $\times$ 71 colors $\times$ 10 iterations) 49,700 ratings.
 
\subsection*{Results and Discussion}

Figure \ref{fig:cors} shows the GPT-4 correlations with human judgments in Experiment 3 for each concept (black bars, labeled ``Multiple Ratings''). The correlations with human ratings for the majority of concepts were positive and statistically reliable, exceeding the critical $r$ value for 61 out of 70 concepts. Although the mean human-GPT correlation across all 70 concepts (mean $r$(69) = .69) was lower than the mean of the human split-half correlation (mean $r$(69) = .93) ($t$(69) = 12.13, $p$<.001), GPT-4 performed better in Experiment 3 than in the previous to experiments. The human-GPT correlations in Experiment 3 were reliably higher than in Experiment 1 (paired $t$(69) = 2.49, p<.05) and Experiment 2 (paired $t$(69) = 3.13, $p$ < .05).

\begin{wrapfigure}{r}{0.4\textwidth}
    \centering
    \includegraphics[width= 0.4\textwidth]{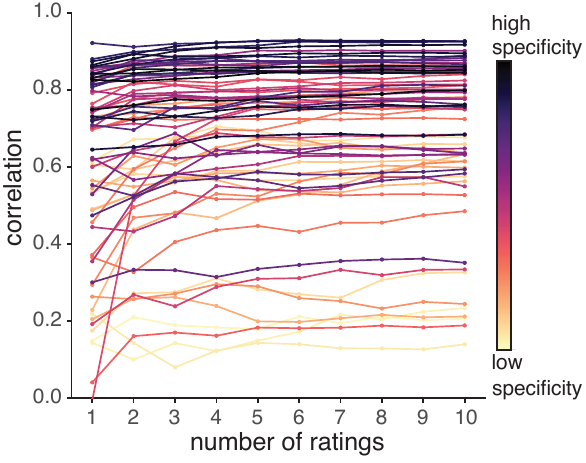}
    \caption{Average correlations between human color-concept association ratings and predicted GPT-4 ratings across all concepts as a function of the number of ratings acquired in Experiment 3. Each line corresponds to a different concept and the color of the line corresponds to the specificity of the concept.}
    \label{fig:num_rate}
    \vspace{-10mm}
\end{wrapfigure}

To investigate the extent to which the human-GPT correlations improve with increased number of ratings, we computed the mean correlation for each concept as a function of the number of ratings used to compute the correlation (0 to 10; Figure \ref{fig:num_rate}). The curves show large concept-to-concept variance, but with correlations generally improving over the first few ratings. There appear to be limited gains, however, following about three ratings per concept.
Mirroring results in Experiments 1-2, specificity ($r$(69) = .51, $p$ < .001) and concreteness ($r$(69) = .29, $p$ < .05) were significantly correlated with the magnitude of human-GPT correlations with the effect of concreteness ($t$ = .01, $p$ = .75) was moderated by the effect of specificity  ($t$ = 16.26, $p$ < .001).

\begin{figure}
    \centering
    \includegraphics[width=1\textwidth]{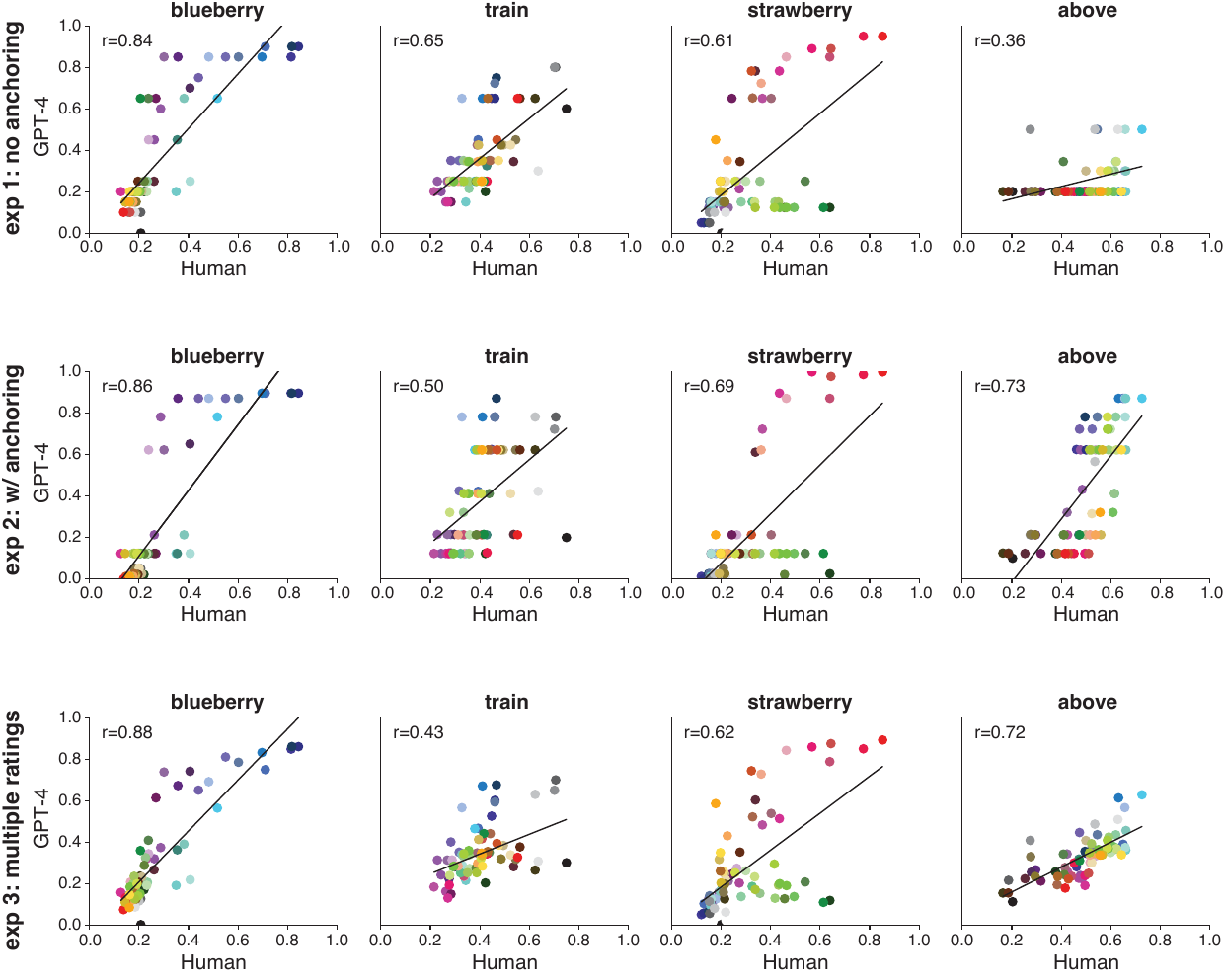}
    \caption{Human color-concept association ratings vs. predicted ratings from GPT-4 across Experiments 1-3 for 4 concepts — blueberry, train, strawberry and above. The effects of anchoring and collecting multiple ratings were on human-GPT-4 alignment varied between concepts.}
    \label{fig:high_med_low}
\end{figure}

These results suggest that acquiring multiple ratings improves performance for estimating color-concept associations, but we note that the margin of improvement was narrow ( mean $r$ = .69 in Experiment 3 vs. mean $r$ = .67 in Experiment 1 the next best performing method). Figure \ref{fig:high_med_low} highlights examples of how no single method is far better than the the others.
On the first row, arranged from left to right we present 4 concepts that show increasing alignment between human and GPT-4 ratings in Experiment 1. 'Above' is the least aligned and 'train' is the most aligned.
However, in Experiment 2, associations for 'above' are better estimated by GPT-4 but associations for 'train' are worse.
And although Experiment 3 \textit{on average} produced the most aligned associations, for 3 of the 4 concepts depicted in Figure \ref{fig:high_med_low} it does not achieve the highest human-GPT correlations relative to results from Experiments 1 and 2.
Meanwhile, associations for concepts like `blueberry' and `strawberry' are approximately equally well captured by all three methods.

\section*{General Discussion}
Motivated by the capabilities of LLM models on a variety of naturalistic tasks \cite{openai2023gpt4,bubeck2023sparks,mirchandani2023large}, we tested GPT-4 on its ability to estimate human-like color-concept associations for a variety of abstract and concrete concepts. 
Across three experiments we found that GPT-4, without any finetuning, performed comparably to existing state-of-the-art methods for estimating color-concept associations \cite{rathore2019estimating,hu2023}. 
Experiment 1 showed that asking GPT-4 for a single rating at zero-temperature for a given color-concept pair yielded association ratings that were significantly correlated with human ratings.
Experiment 2 showed that adding an anchoring task similar to task-prefixing did not improve GPT-4's performance relative to Experiment 1. Experiment 3 showed that correlations with human ratings improved slightly but significantly when model responses were sampled stochastically from token output distributions and averaged.

While details regarding the precise nature of multi-modal pretraining that GPT-4 has undergone are unclear, the human-like color-concept associations the model generates when probed using language alone suggests that it acquires relatively rich visual associations accessible from language input \cite{kim2021shared,xu2023does,hansen2022semantic, suresh2023semantic, lupyan2020effects}.
The fact that GPT-4 showed good performance even for highly abstract concepts such as `night' and `love,' suggest that it is not challenged by abstractness \emph{per se}. 
Indeed, across all three experiments we showed that GPT-4's ability to estimate color-concept associations for a given concept was predicted by the specificity of the true underlying association distribution. Critically, this effect of specificity completely moderated the effect of a concept's concreteness indicating that the key factor determining the model's ability to estimate associations was specificity and \textit{not} whether a concept is abstract or concrete.
The results suggest that GPT-4, and performant LLMs in general, can be used as a tool for automatically estimating color-concept associations for a very broad range of concepts, including highly abstract concepts that do not have directly observable colors. 

One limitation of our approach is that our human associations rating data were collected from a university undergraduate population in the United States and our GPT-4 rating task was only performed in English. Future work should investigate whether GPT-4 can estimate association ratings between colors and concepts in other languages and test whether those ratings align with human ratings collected from native speakers of those languages. 
A second limitation is that hex codes are a coarse proxy for probing knowledge about perceptual stimuli such as color and do not perfectly align with the human task where participants were shown colored squares. 
With the advent of vision-language models like LLaVA \cite{li2023llava} and GPT-4 Vision that can take images in addition to text as input, future work can seek to test whether human-like color-concept associations can be estimated by providing the same colored patches to the model as are presented to human participants.

We began this paper posing the question of how color-concept associations for abstract concepts could be acquired for concepts that do not clearly encompass concrete, perceptible items in the world. 
The present results suggest that such associations are latent in the statistical structure of natural language and color information that LLMs like GPT-4 are able to capture and reproduce (using hexadecimal values).
To the extent that humans are also sensitive to this statistical structure in their linguistic and perceptual experiences, LLMs constitute a strong family of candidate models for how humans might acquire color-concept associations.
To validate these models as models of human learning, future behavioral experiments will need to carefully manipulate the structure and frequency of color-concept co-occurrences and test the effect of varied statistical co-occurrences on learned associations.

The current effectiveness of our approach is particularly relevant to the field of information visualization, where the color that is easiest to interpret for a concept in a visualization is sometimes \textit{not} the most associated color (see Schloss et al. \cite{schloss2018} for a detailed overview).
It is therefore important to quantify association ratings between concepts and many colors, sampled over all of perceptual color space when designing effective visualizations.
We show that LLMs can be used to automate this process effectively.

Taken together, our study presents an easy-to-implement method for automatically estimating color-concept associations using the knowledge latent in LLMs. 
Our results highlight that a learning system that is exposed to concepts in natural language in addition to functional proxies for perceptual information (like SVG strings) and some degree of multimodal training using images is able to produce human-like associations between concepts and highly specific percepts, colors that span perceptual color space.
GPT-4, while largely opaque to investigation, thus provides an existence proof of a model that can learn structured associations between concepts of varying degrees of abstraction and perceptual properties.

\section*{Acknowledgements}
The authors thank Melissa Schoenlein, Clementine Zimnicki, and members of the Schloss Visual Reasoning Lab for helpful feedback on this project and Robert Hawkins for their help with the GPT rating experiment. This material is based upon work supported by the National Science Foundation under Grant No. NSF BCS-1945303 to KBS. Any opinions, findings, and conclusions or recommendations expressed in this material are those of the author(s) and do not necessarily reflect the views of the National Science Foundation.

\bibliographystyle{unsrt}  
\bibliography{references}  
\newpage 
\appendix 
\counterwithin{figure}{section}
\counterwithin{table}{section}

\section{Modeling color concept associations using color space regression models}

\begin{figure}[htb!!]
    \centering
    \includegraphics[width=1\textwidth]{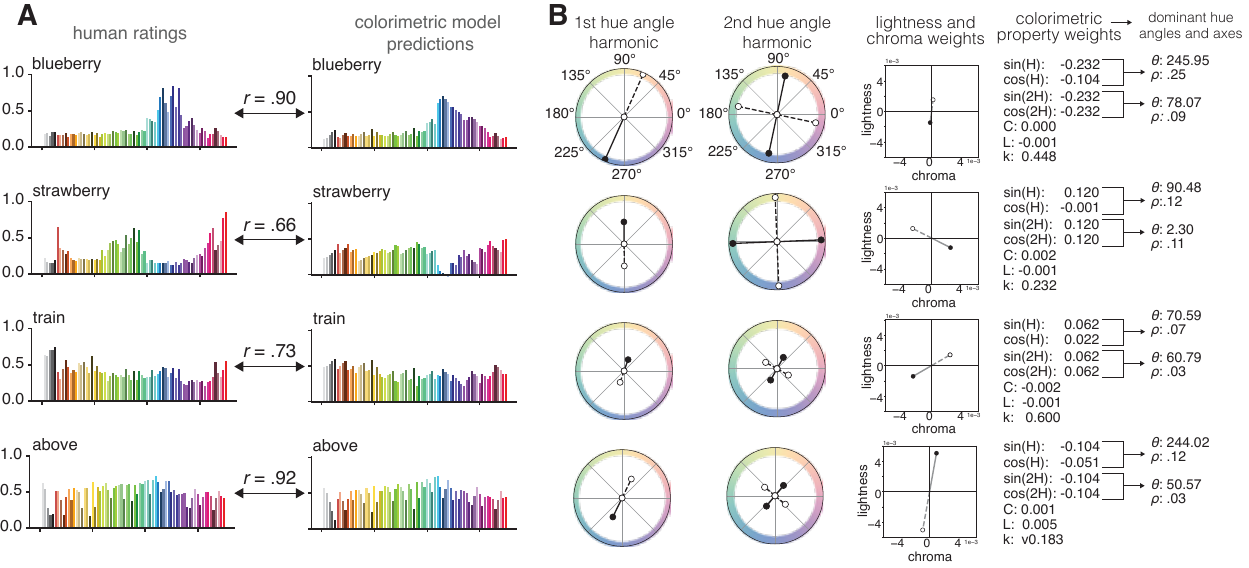}
    \caption{(A) Mean human color-concept association ratings between the concept and the \texttt{UW-71} colors and predicted associations from color space regression models. (B) Visualization of the coefficients of the color space regression models. From left to right, dominant hue angle, dominant hue axis, lightness and chroma, and numeric representations of the coefficients. The weights on the sines and cosines of the hue angle and the double hue angle are used to compute the dominant hues and axes visualized in the middle columns. }
    \label{fig:colorimetric}
    
\end{figure}

In this paper our goal was to test whether LLMs like GPT-4 can produce human-like color-concept association ratings.
In this context, it is helpful to characterize human color-concept associations. 
One way to do this is to compute metrics on the distribution of associations for a given concept such as \textit{specificity}.
Another way, that provides detailed information about a concept's associations in terms of how they relate to color space properties is to use color space (colorimetric) regression models that use coordinates in color space to explain patterns of data sampled across color space.
These models were first developed and tested to model human color preferences \cite{hurlbert2007biological, ling2007new}, but have recently been extended to model human color-concept association ratings (refer to supplementary materials in Schoenlein et al. (2022) \cite{schoenlein2022unifying} for more details). 
Schloss et al. \cite{schloss2018modeling} showed that a cylindrical model in CIELCh space with two harmonics best explained color preference data and this same model was subsequently used by Schoenlein et al. (2022)  \cite{schoenlein2022unifying} to model people's color-concept associations.
The model consists of seven predictors —  lightness (L), chroma (C), the first harmonic of hue angle (sin(h)
and cos(h)), the second harmonic of hue angle (sin(2h) and cos(2h)), and
a constant (k).
These models can be use to effectively estimate human color-concept associations (measured via model fit) and investigating the coefficients or weights of the models highlight the relationship between color space properties and association ratings on a concept-by-concept basis.
Separate models were fit for each concept where the predictors were the 6 colorimetric properties of each of the \texttt{UW-71} colors plus the constant and the predicted value was the association rating between the concept and that color. 
Models were fit using ordinary least squares (OLS). The quality of the model fits were assessed by computing the correlation between true ratings and predicted ratings using the fitted coefficients.
Figure \ref{fig:colorimetric} shows estimated associations and visual depictions of model weights for four models fit to four difference concepts. Refer to table \ref{tab:reg_coefs} for coefficients for all concepts.

{
\captionsetup{skip=10pt} 
\begin{longtable}{l|r|r|r|r|r|r|r}
\midrule
\textbf{Concept} & \textbf{L} & \textbf{C} &\textbf{ cos(H)} & \textbf{sin(H)} & \textbf{cos(2H)} & \textbf{sin(2H) }& \textbf{k}\\
\midrule
above & 0.005 & 0.001 & -0.051 & -0.104 & -0.104 & -0.104 & 0.183 \\
\midrule
angry & -0.005 & 0.001 & 0.063 & 0.099 & 0.099 & 0.099 & 0.526 \\
\midrule
apple & -0.001 & 0.003 & -0.026 & 0.201 & 0.201 & 0.201 & 0.151 \\
\midrule
banana & 0.002 & -0.0 & 0.027 & 0.177 & 0.177 & 0.177 & 0.062 \\
\midrule
beach & 0.004 & -0.001 & -0.106 & -0.146 & -0.146 & -0.146 & 0.292 \\
\midrule
bear & -0.003 & -0.004 & 0.048 & 0.167 & 0.167 & 0.167 & 0.656 \\
\midrule
below & -0.005 & -0.002 & -0.021 & 0.039 & 0.039 & 0.039 & 0.808 \\
\midrule
beside & 0.001 & -0.0 & -0.007 & -0.027 & -0.027 & -0.027 & 0.431 \\
\midrule
bird & 0.0 & -0.002 & 0.003 & 0.023 & 0.023 & 0.023 & 0.553 \\
\midrule
blizzard & 0.003 & -0.004 & -0.062 & -0.272 & -0.272 & -0.272 & 0.422 \\
\midrule
blueberry & -0.001 & -0.0 & -0.104 & -0.232 & -0.232 & -0.232 & 0.448 \\
\midrule
boat & 0.001 & -0.004 & -0.038 & -0.159 & -0.159 & -0.159 & 0.513 \\
\midrule
car & -0.0 & -0.003 & 0.05 & -0.009 & -0.009 & -0.009 & 0.485 \\
\midrule
carrot & -0.001 & 0.002 & -0.026 & 0.209 & 0.209 & 0.209 & 0.22 \\
\midrule
celery & -0.0 & 0.0 & -0.155 & 0.19 & 0.19 & 0.19 & 0.182 \\
\midrule
cherry & -0.002 & 0.003 & 0.057 & 0.093 & 0.093 & 0.093 & 0.179 \\
\midrule
comfort & 0.003 & -0.002 & 0.022 & -0.07 & -0.07 & -0.07 & 0.43 \\
\midrule
corn & 0.001 & -0.0 & 0.0 & 0.206 & 0.206 & 0.206 & 0.103 \\
\midrule
dawn & 0.003 & -0.001 & 0.054 & -0.033 & -0.033 & -0.033 & 0.332 \\
\midrule
day & 0.007 & 0.001 & -0.068 & -0.083 & -0.083 & -0.083 & 0.087 \\
\midrule
disgust & -0.004 & -0.001 & -0.04 & 0.201 & 0.201 & 0.201 & 0.652 \\
\midrule
dress & 0.002 & -0.001 & 0.077 & -0.112 & -0.112 & -0.112 & 0.393 \\
\midrule
driving & -0.0 & -0.0 & -0.041 & 0.041 & 0.041 & 0.041 & 0.475 \\
\midrule
drought & -0.0 & -0.003 & 0.11 & 0.193 & 0.193 & 0.193 & 0.469 \\
\midrule
dusk & -0.005 & -0.002 & 0.039 & -0.051 & -0.051 & -0.051 & 0.815 \\
\midrule
eating & 0.0 & 0.001 & -0.005 & 0.102 & 0.102 & 0.102 & 0.344 \\
\midrule
efficiency & 0.003 & 0.001 & -0.033 & -0.055 & -0.055 & -0.055 & 0.266 \\
\midrule
eggplant & -0.002 & 0.0 & 0.062 & 0.019 & 0.019 & 0.019 & 0.332 \\
\midrule
evil & -0.006 & -0.0 & 0.006 & 0.084 & 0.084 & 0.084 & 0.699 \\
\midrule
far & -0.002 & -0.002 & -0.013 & -0.014 & -0.014 & -0.014 & 0.676 \\
\midrule
fearful & -0.004 & -0.001 & 0.022 & 0.046 & 0.046 & 0.046 & 0.63 \\
\midrule
field & -0.001 & -0.0 & -0.079 & 0.175 & 0.175 & 0.175 & 0.406 \\
\midrule
fish & -0.0 & -0.001 & -0.087 & -0.071 & -0.071 & -0.071 & 0.548 \\
\midrule
frog & -0.003 & -0.001 & -0.179 & 0.166 & 0.166 & 0.166 & 0.547 \\
\midrule
grape & -0.002 & 0.002 & 0.034 & 0.03 & 0.03 & 0.03 & 0.268 \\
\midrule
greed & -0.004 & -0.0 & -0.05 & 0.133 & 0.133 & 0.133 & 0.633 \\
\midrule
happy & 0.006 & 0.004 & -0.006 & -0.091 & -0.091 & -0.091 & -0.015 \\
\midrule
hurricane & -0.002 & -0.002 & -0.118 & -0.128 & -0.128 & -0.128 & 0.675 \\
\midrule
justice & 0.001 & -0.002 & 0.049 & -0.056 & -0.056 & -0.056 & 0.441 \\
\midrule
leisure & 0.002 & -0.0 & -0.045 & -0.092 & -0.092 & -0.092 & 0.42 \\
\midrule
lemon & 0.003 & -0.0 & 0.036 & 0.115 & 0.115 & 0.115 & 0.051 \\
\midrule
lightning & 0.004 & -0.002 & 0.048 & -0.054 & -0.054 & -0.054 & 0.202 \\
\midrule
lion & -0.0 & -0.001 & 0.102 & 0.212 & 0.212 & 0.212 & 0.389 \\
\midrule
love & 0.003 & 0.003 & 0.167 & -0.075 & -0.075 & -0.075 & 0.069 \\
\midrule
mango & 0.001 & 0.002 & 0.029 & 0.164 & 0.164 & 0.164 & 0.132 \\
\midrule
mushroom & 0.0 & -0.004 & 0.104 & 0.17 & 0.17 & 0.17 & 0.387 \\
\midrule
near & 0.002 & 0.001 & 0.007 & -0.025 & -0.025 & -0.025 & 0.304 \\
\midrule
night & -0.007 & -0.003 & -0.013 & -0.053 & -0.053 & -0.053 & 0.923 \\
\midrule
noon & 0.006 & 0.001 & -0.037 & -0.056 & -0.056 & -0.056 & 0.1 \\
\midrule
ocean & 0.0 & -0.001 & -0.247 & -0.29 & -0.29 & -0.29 & 0.514 \\
\midrule
pants & -0.001 & -0.005 & 0.024 & -0.046 & -0.046 & -0.046 & 0.622 \\
\midrule
peace & 0.006 & -0.001 & 0.018 & -0.102 & -0.102 & -0.102 & 0.221 \\
\midrule
peach & 0.002 & 0.001 & 0.126 & 0.125 & 0.125 & 0.125 & 0.073 \\
\midrule
plane & 0.002 & -0.004 & 0.01 & -0.113 & -0.113 & -0.113 & 0.408 \\
\midrule
reliability & 0.002 & -0.001 & 0.025 & -0.035 & -0.035 & -0.035 & 0.361 \\
\midrule
sad & -0.004 & -0.004 & -0.058 & -0.069 & -0.069 & -0.069 & 0.856 \\
\midrule
safety & 0.003 & -0.001 & -0.014 & -0.048 & -0.048 & -0.048 & 0.403 \\
\midrule
sandstorm & 0.001 & -0.003 & 0.148 & 0.193 & 0.193 & 0.193 & 0.343 \\
\midrule
shirt & 0.001 & -0.003 & 0.038 & -0.059 & -0.059 & -0.059 & 0.614 \\
\midrule
shoes & 0.001 & -0.004 & 0.082 & -0.028 & -0.028 & -0.028 & 0.51 \\
\midrule
sky & 0.002 & -0.002 & -0.152 & -0.303 & -0.303 & -0.303 & 0.439 \\
\midrule
sleeping & -0.003 & -0.004 & -0.026 & -0.117 & -0.117 & -0.117 & 0.863 \\
\midrule
socks & 0.001 & -0.004 & 0.086 & 0.001 & 0.001 & 0.001 & 0.465 \\
\midrule
speed & 0.003 & 0.004 & -0.021 & -0.007 & -0.007 & -0.007 & 0.083 \\
\midrule
strawberry & -0.001 & 0.002 & -0.001 & 0.12 & 0.12 & 0.12 & 0.232 \\
\midrule
sunset & 0.002 & 0.003 & 0.132 & -0.012 & -0.012 & -0.012 & 0.137 \\
\midrule
train & -0.001 & -0.002 & 0.022 & 0.062 & 0.062 & 0.062 & 0.6 \\
\midrule
truck & -0.002 & -0.004 & 0.041 & 0.029 & 0.029 & 0.029 & 0.593 \\
\midrule
watermelon & -0.001 & 0.003 & -0.06 & 0.167 & 0.167 & 0.167 & 0.233 \\
\midrule
working & -0.001 & -0.003 & -0.017 & 0.043 & 0.043 & 0.043 & 0.639 \\
\bottomrule

\caption{ Regression coefficients for color space regression models fit to each concept's color-concept associations. L refers to lightness, C refers to chroma, H refers to hue angle and k is a constant.}
\label{tab:reg_coefs}
\end{longtable}
}


\section{Human and GPT-4 Color-Concept Association Distributions}

\begin{figure}[htb!]
    \centering
    \includegraphics[width=.9\textwidth]{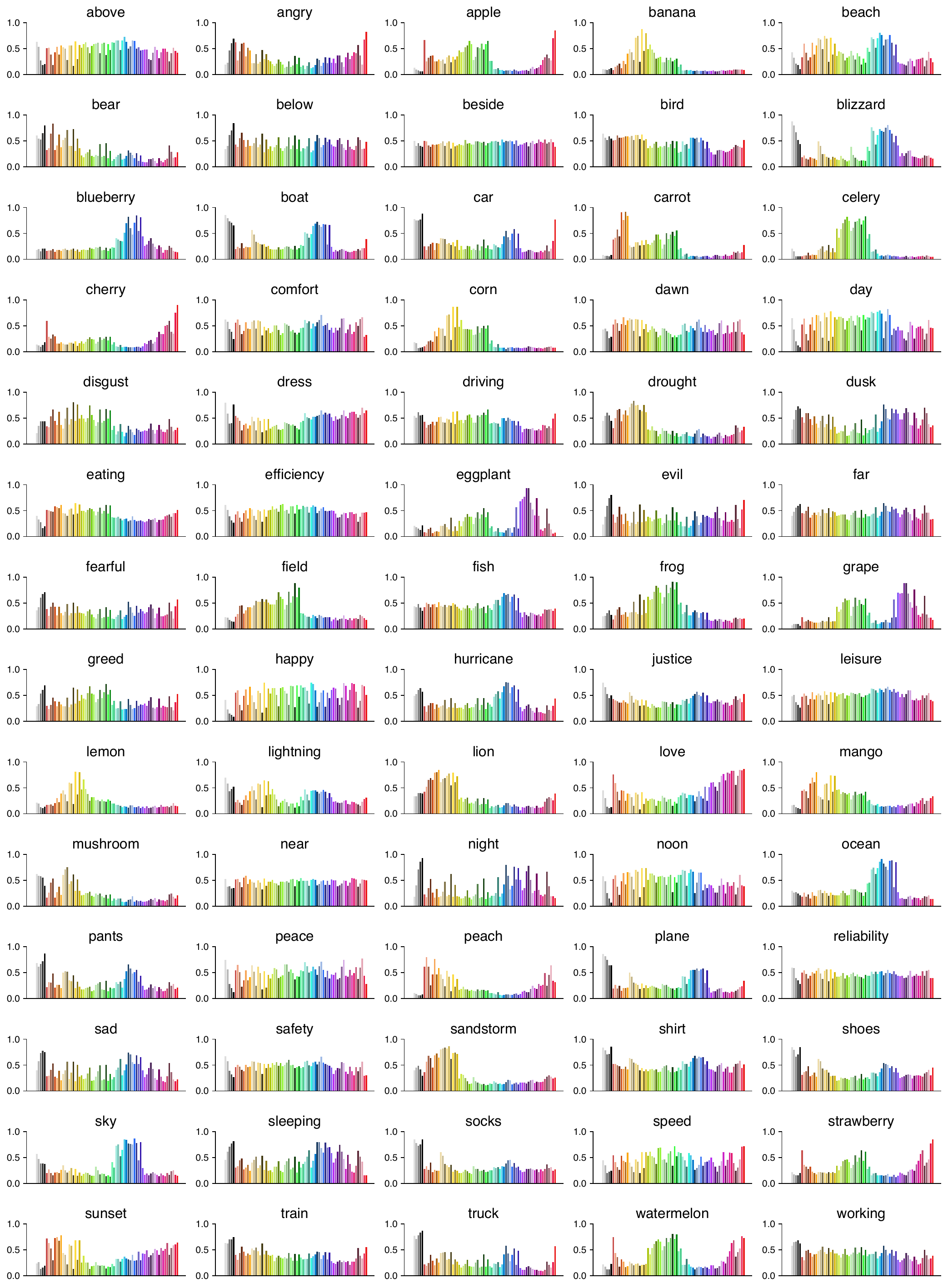}
    \caption{Mean \textbf{human} color-concept association ratings for all 70 concepts across the \texttt{UW-71} colors.}
    \label{fig:enter-label}
\end{figure}

\begin{figure}
    \centering
    \includegraphics[width=.9\textwidth]{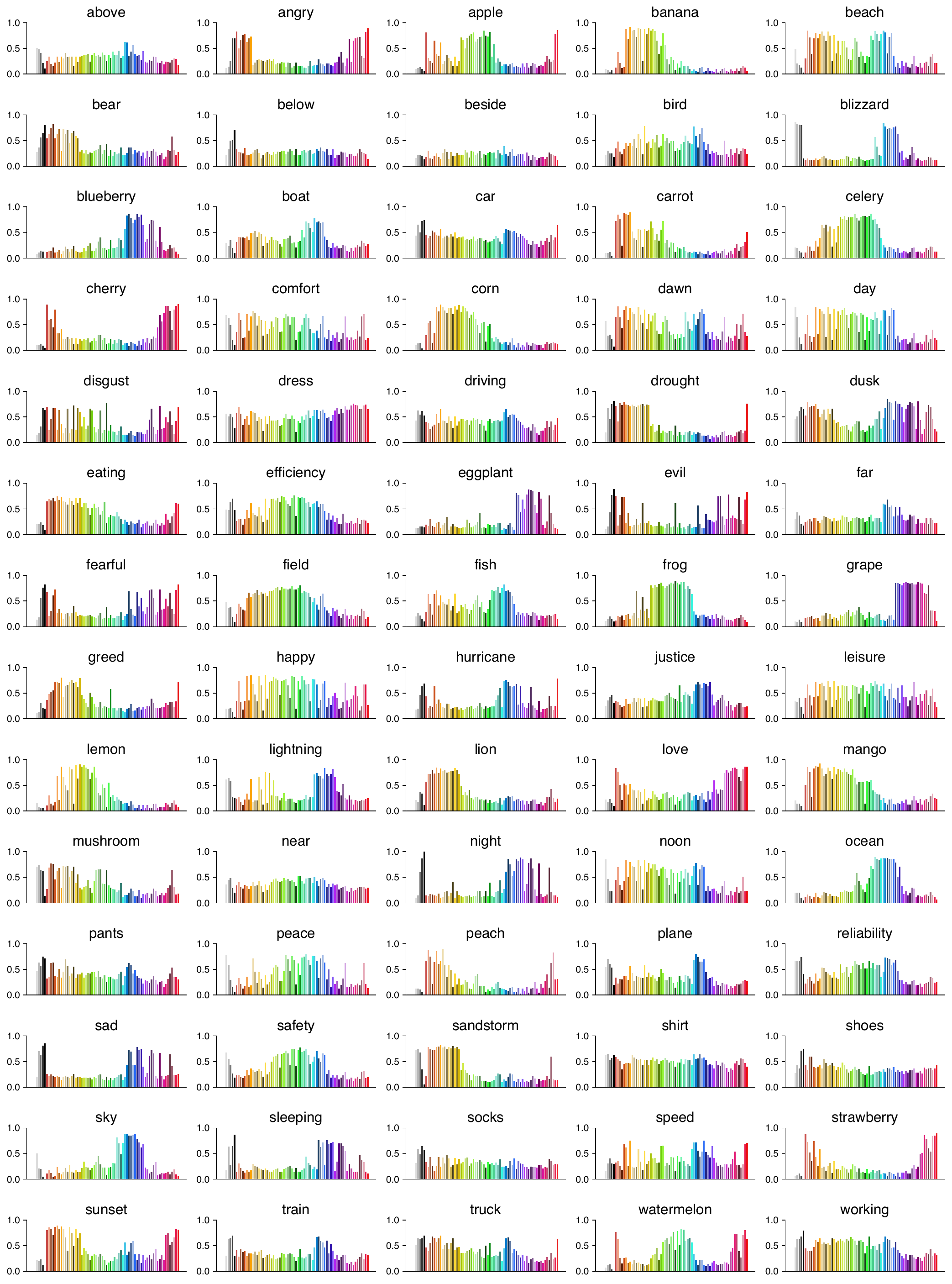}
    \caption{\textbf{GPT-4} color concept associations from \textbf{Experiment 1} for all 70 concepts across the \texttt{UW-71} colors..}
    \label{fig:enter-label}
\end{figure}

\begin{figure}
    \centering
    \includegraphics[width=.9\textwidth]{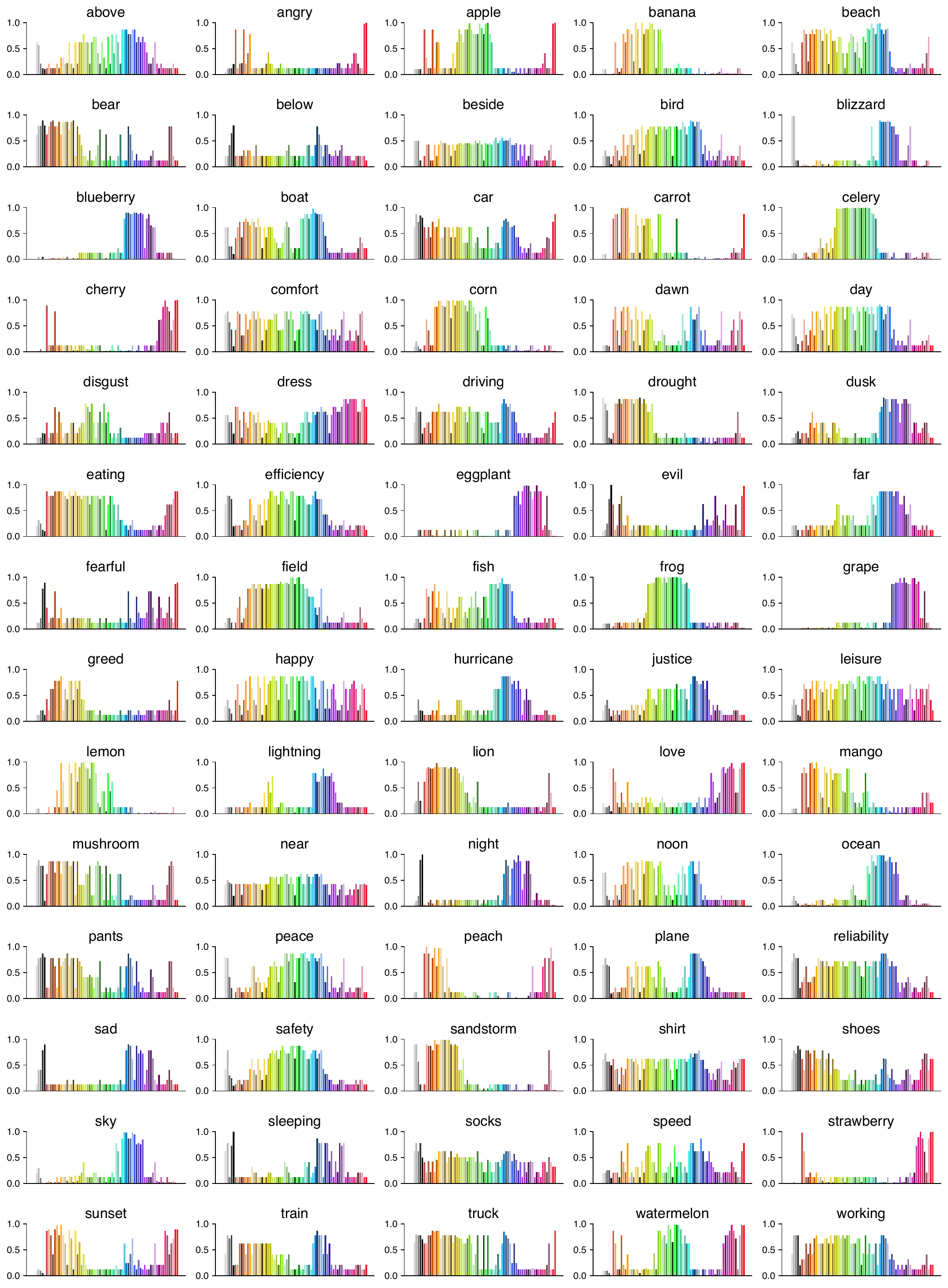}
     \caption{\textbf{GPT-4} color concept associations from \textbf{Experiment 2} for all 70 concepts across the \texttt{UW-71} colors..}
    \label{fig:enter-label}
\end{figure}

\begin{figure}
    \centering
    \includegraphics[width=.9\textwidth]{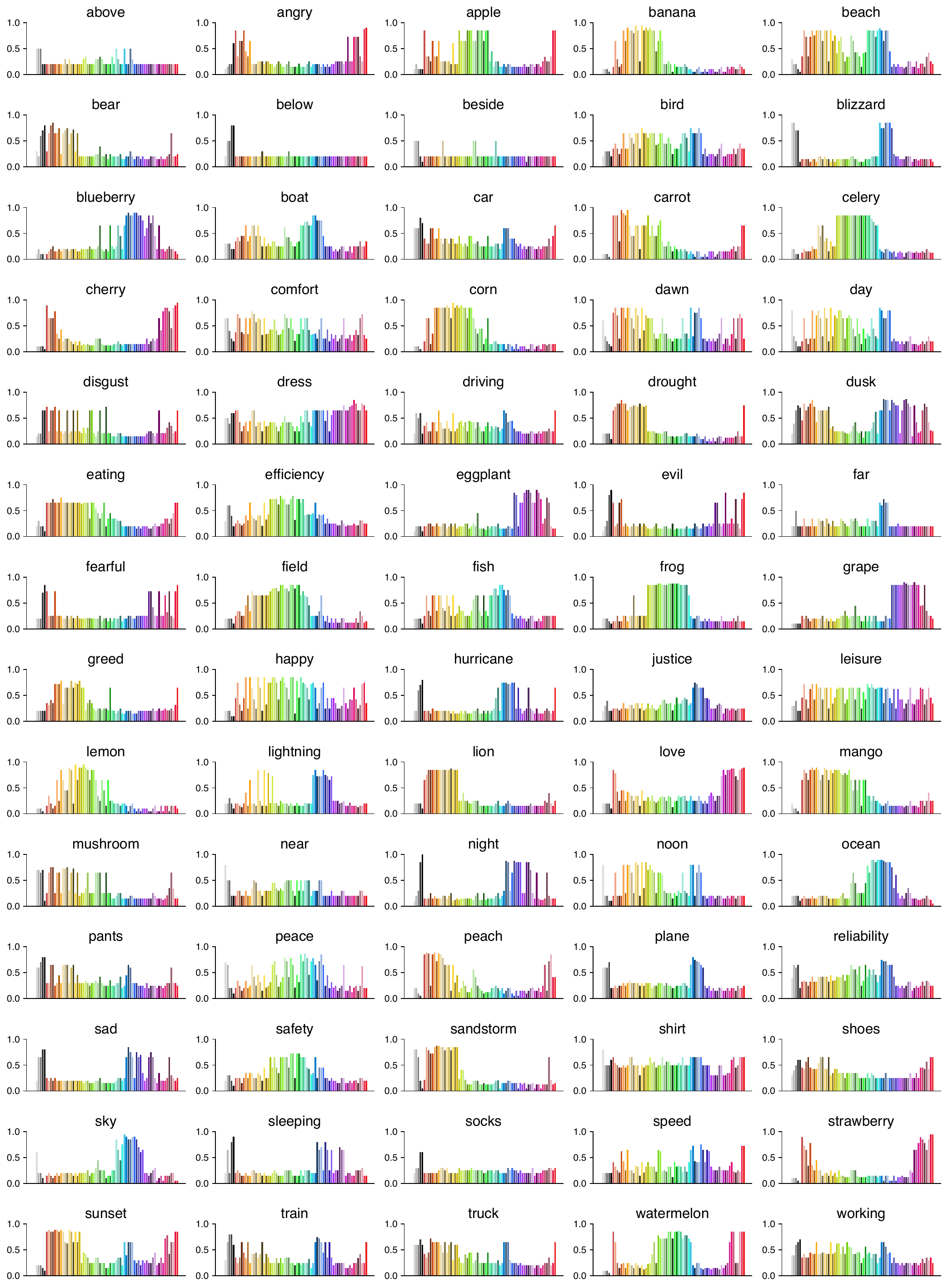}
       \caption{\textbf{GPT-4} color concept associations from \textbf{Experiment 3} for all 70 concepts across the \texttt{UW-71} colors.}
    \label{fig:enter-label}
\end{figure}

\begin{figure}
       \caption{Correlations between human color-concept association ratings and estimated associations from GPT-4 for Experiments 1,2, and 3.}
    \centering
    \includegraphics[width=.55\textwidth]{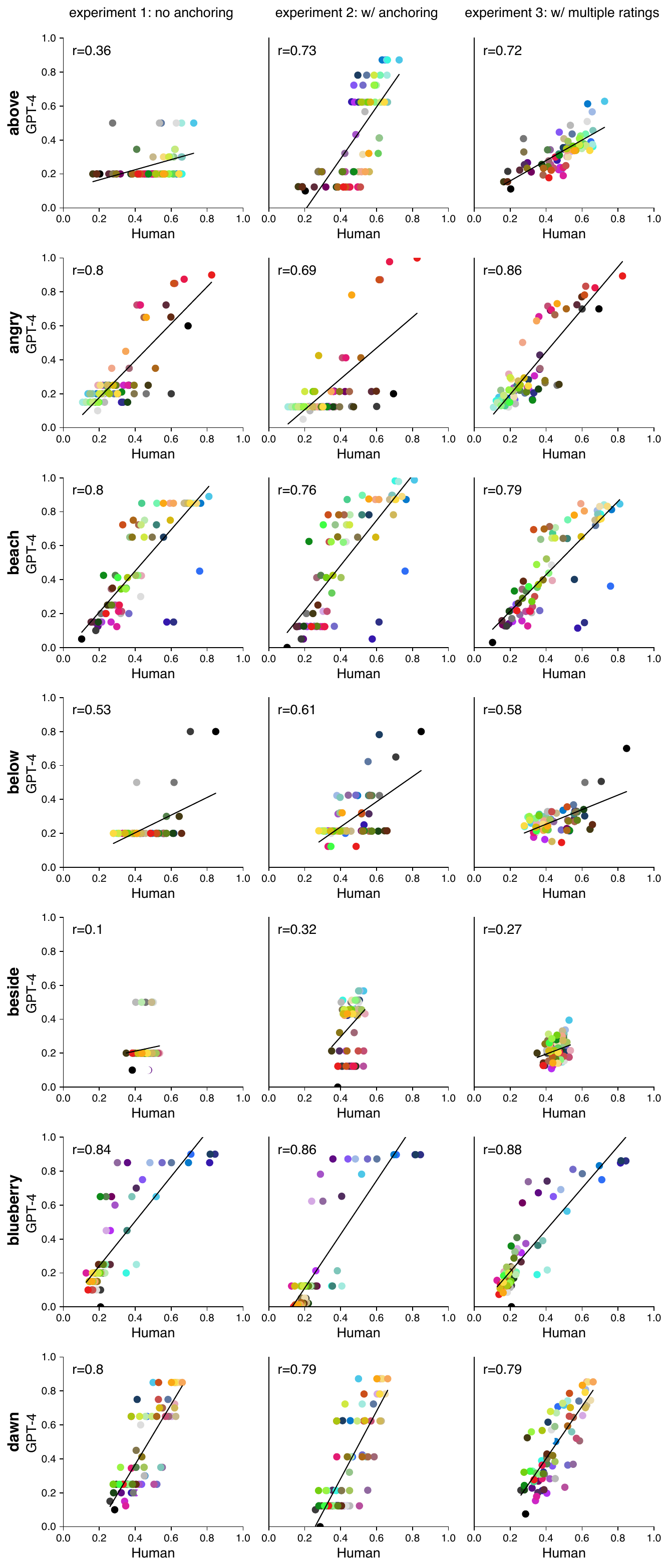}

\end{figure}

\begin{figure}
    \centering
    \includegraphics[width=.55\textwidth]{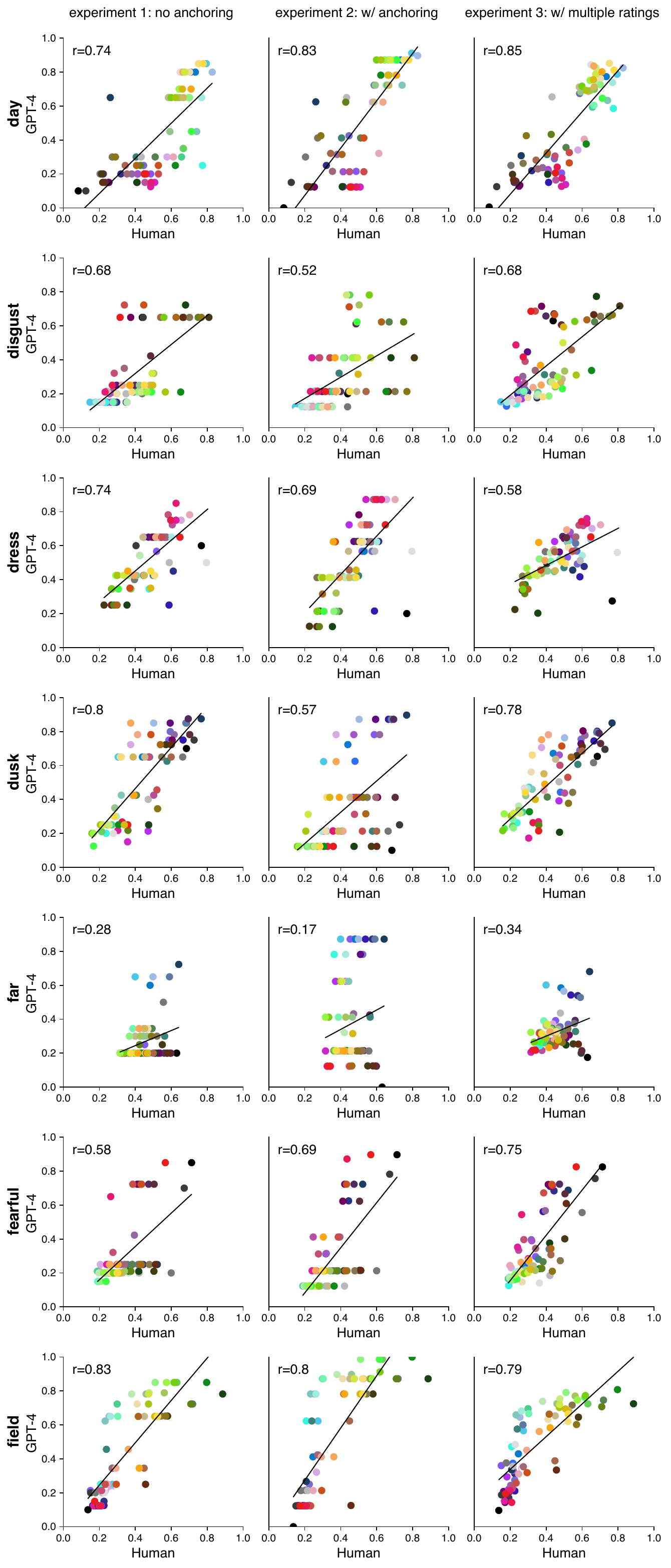}

\end{figure}

\begin{figure}
    \centering
    \includegraphics[width=.55\textwidth]{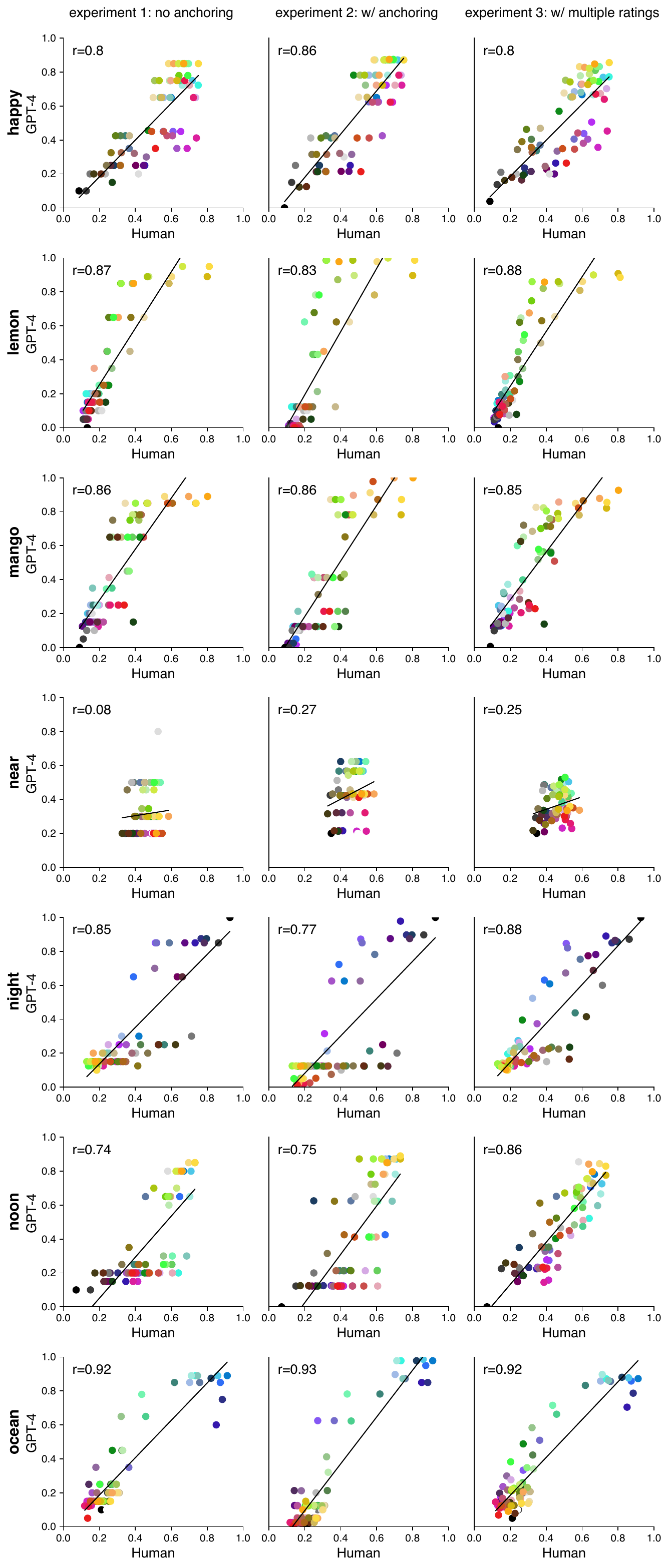}

\end{figure}

\begin{figure}
    \centering
    \includegraphics[width=.55\textwidth]{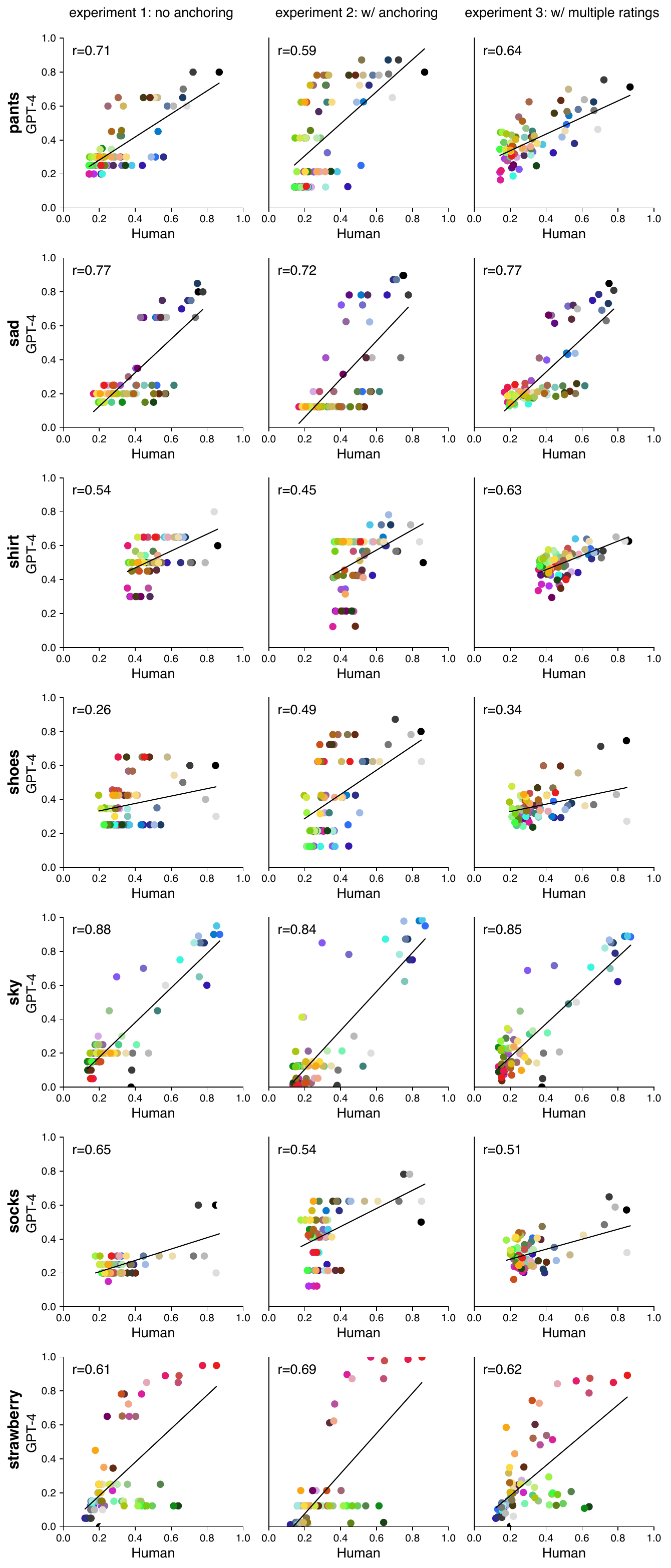}

\end{figure}

\begin{figure}
    \centering
    \includegraphics[width=.55\textwidth]{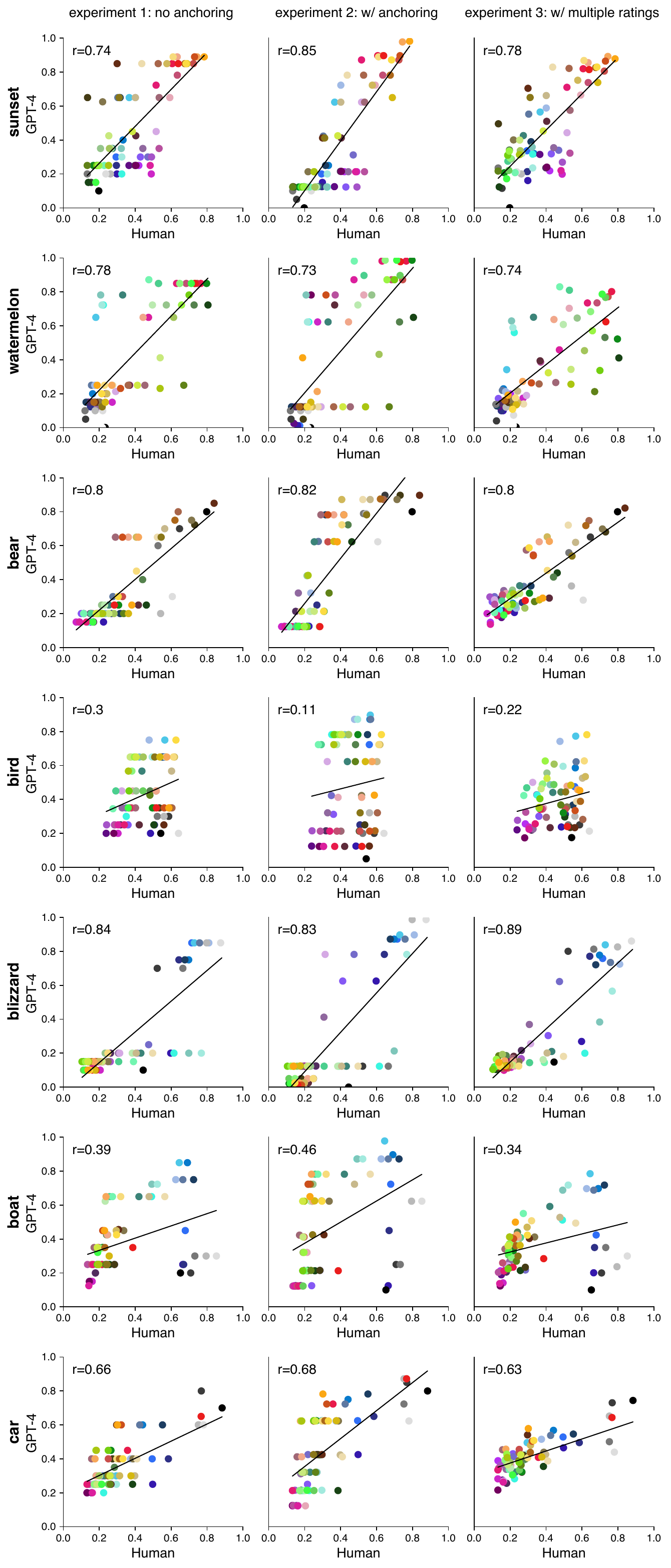}
     
\end{figure}

\begin{figure}
    \centering
    \includegraphics[width=.55\textwidth]{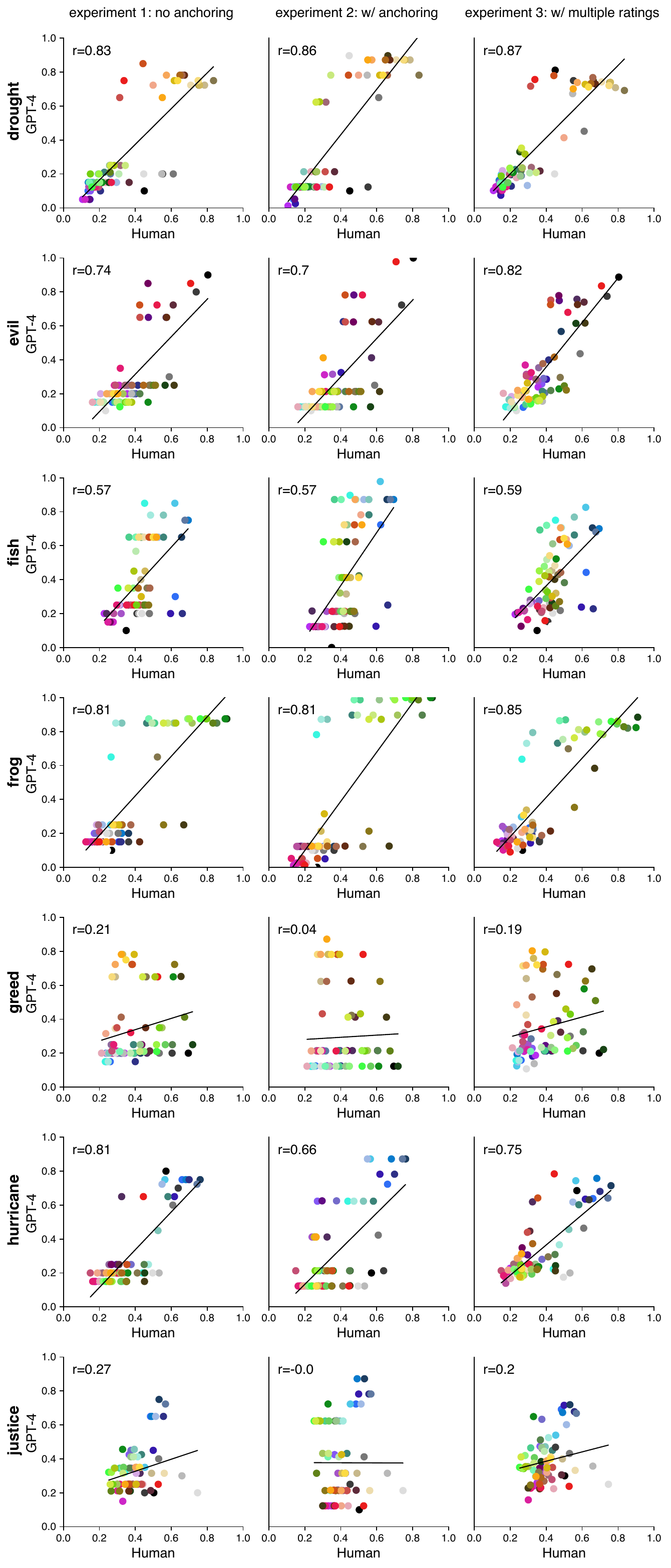}
  
\end{figure}

\begin{figure}
    \centering
    \includegraphics[width=.55\textwidth]{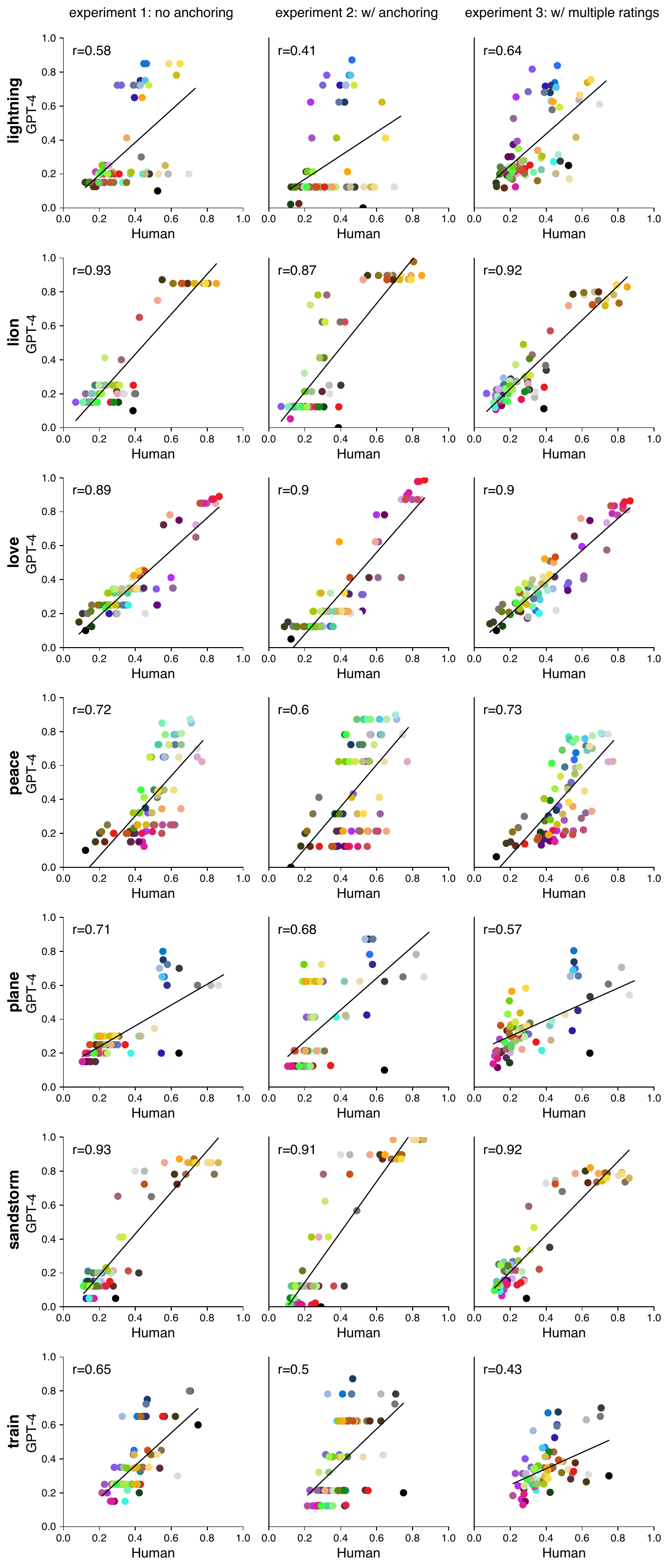}

\end{figure}

\begin{figure}
    \centering
    \includegraphics[width=.55\textwidth]{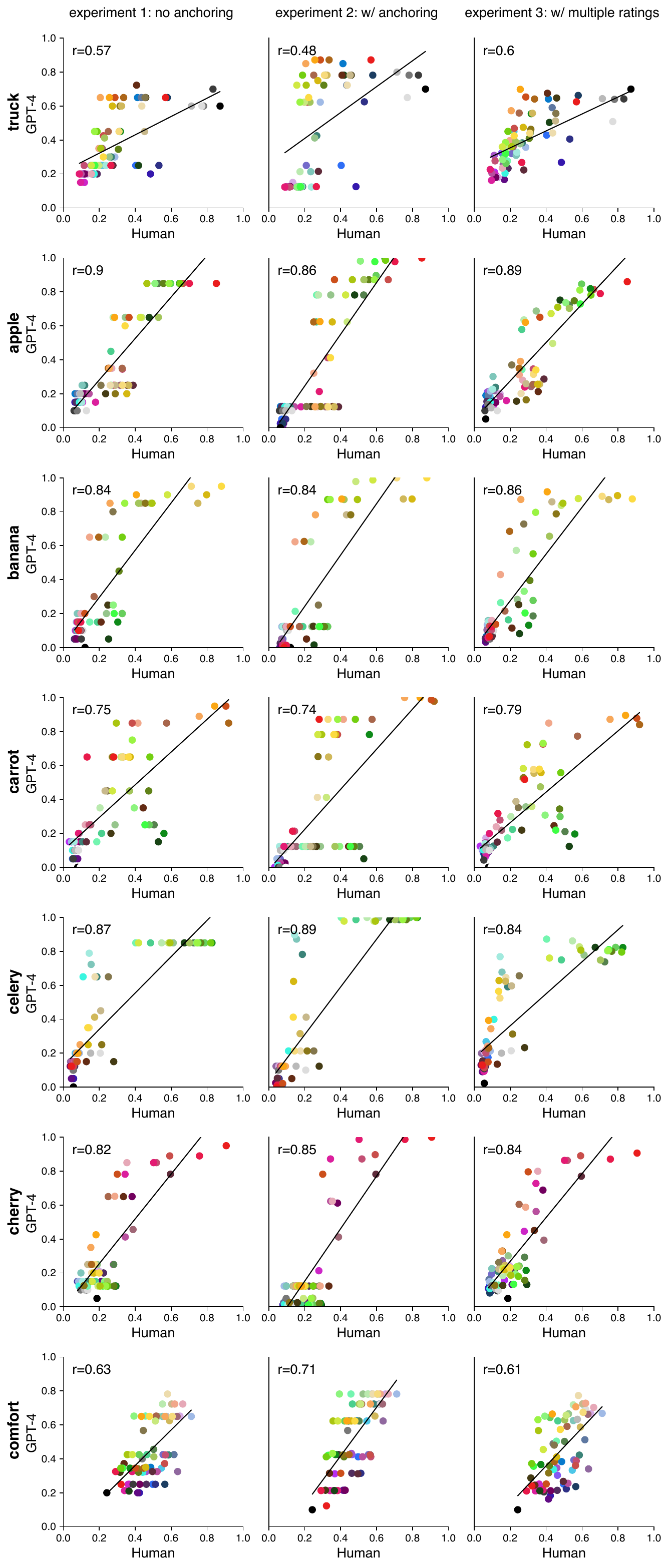}

\end{figure}

\begin{figure}
    \centering
    \includegraphics[width=.55\textwidth]{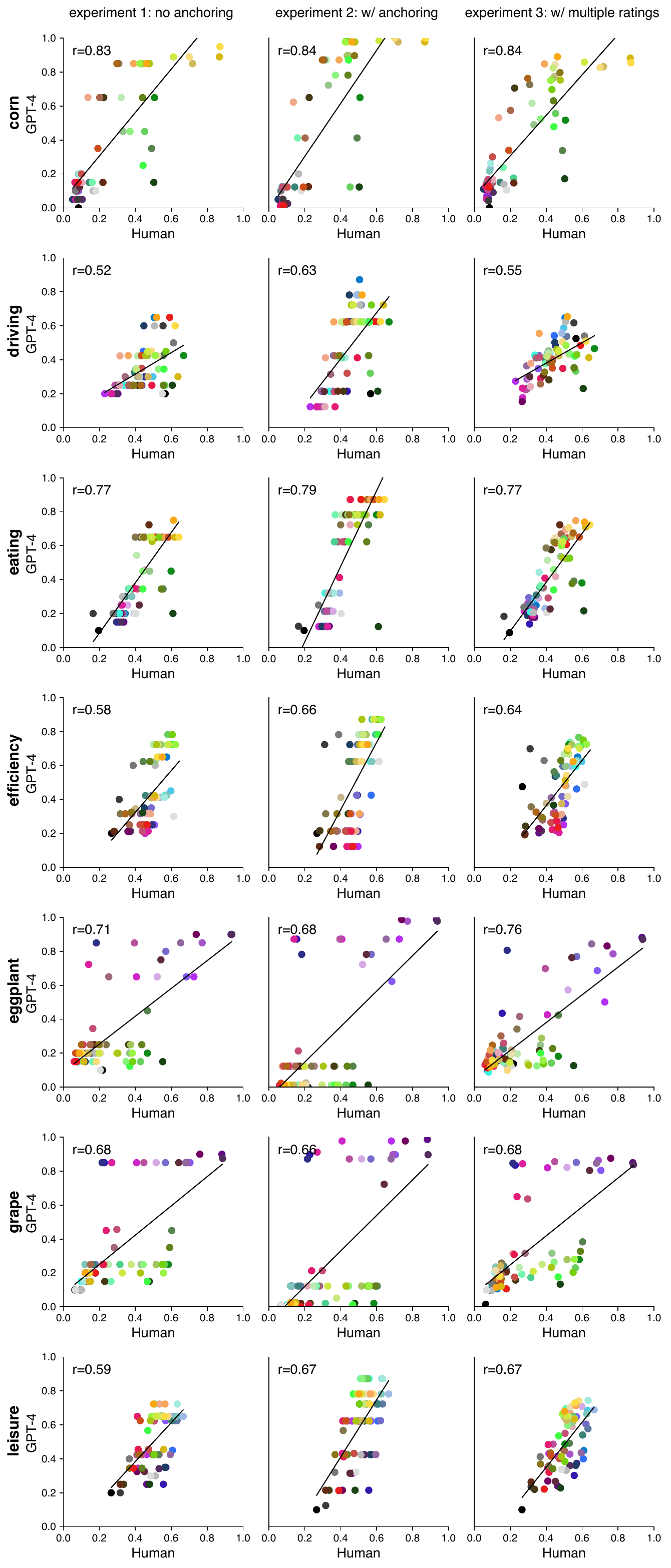}
\end{figure}

\begin{figure}
    \centering
    \includegraphics[width=.55\textwidth]{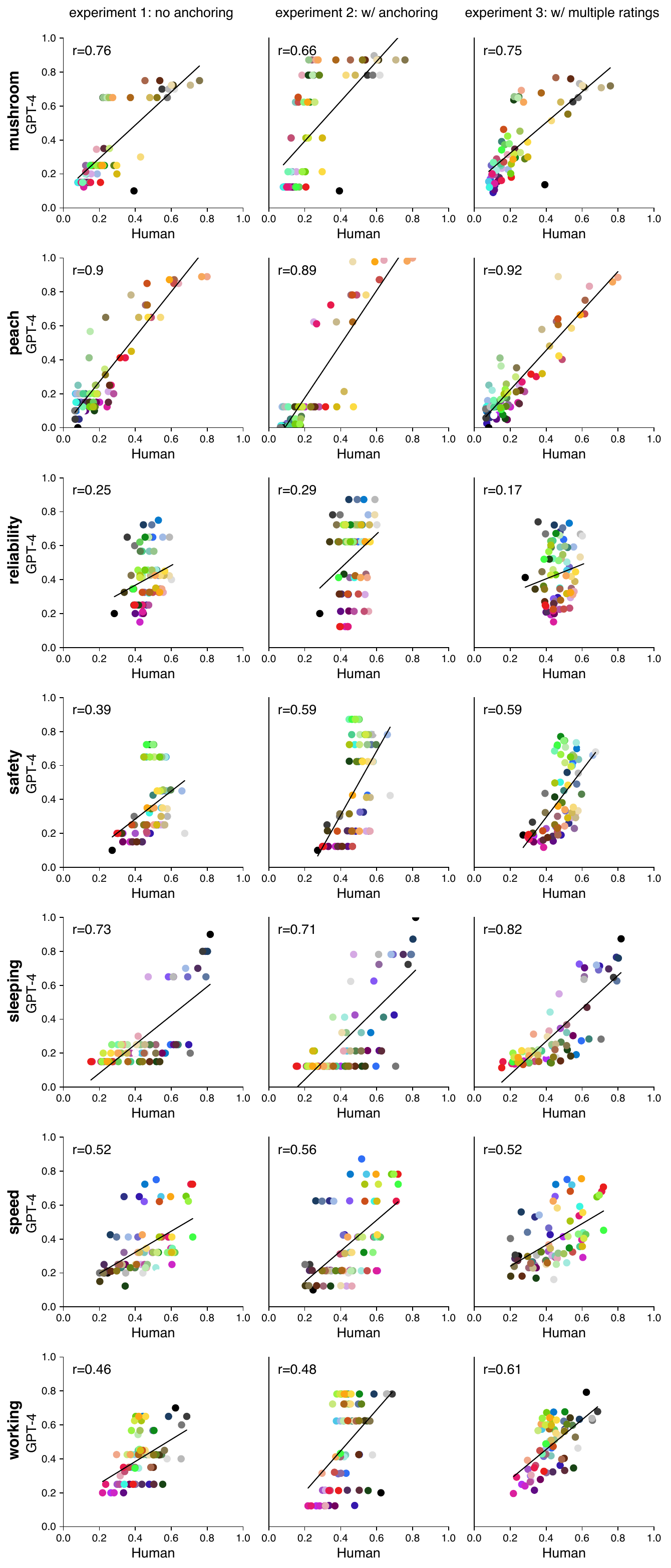}

\end{figure}

{
\captionsetup{skip=10pt} 
\sisetup{round-mode=places}
\begin{longtable}
{c*{1}{
    |S[round-precision=0]
    |S[round-precision=3]
    |S[round-precision=3]
    |S[round-precision=2]
    |S[round-precision=2]
    |S[round-precision=2]
    |S[round-precision=2]
}}
\midrule
\textbf{Color} & \textbf{Sorted Position} & \textbf{x} &\textbf{y} & \textbf{Y} & \textbf{L*} & \textbf{a*}& \textbf{b*}\\
\midrule
1 & 50 & 0.17813 & 0.14021 & 18.419 & 50 & 28.891 & -73.589 \\
\midrule
2 &53 & 0.1742 & 0.082514 & 4.4155 & 25 & 53.857 & -72.28 \\
\midrule
3 & 54 & 0.21726 & 0.13588 & 18.419 & 50 & 53.857 & -72.28 \\
\midrule
4 &55& 0.2591 & 0.13088 & 18.419 & 50 & 78.822 & -70.972\\
\midrule
5 &46& 0.18715 & 0.19157 & 18.419 & 50 & 2.6168 & -49.931\\
\midrule
6 &51& 0.19063 & 0.1298 & 4.4155 & 25 & 27.583 & -48.623\\
\midrule
7&52 & 0.23145 & 0.18448 & 8.419 & 50 & 27.583 & -48.623\\
\midrule
8&57 & 0.25495 & 0.12284 & 4.4155 & 25 & 52.548 & -47.315\\
\midrule
9 &56 & 0.27872 & 0.17635 & 18.419 & 50 & 52.548 & -47.315\\
\midrule
10&61  & 0.32783 & 0.1674 & 18.419 & 50 & 77.514 & -46.006\\
\midrule
11&45  & 0.22397 & 0.28399 & 48.278 & 75 & -23.657 & -26.274\\
\midrule
12&47  & 0.2081 & 0.21415 & 4.4155 & 25 & 1.3084 & -24.966\\
\midrule
13&48  & 0.24471 & 0.25395 & 18.419 & 50 & 1.3084 & -24.966\\
\midrule
14&49  & 0.26261 & 0.27354 & 48.278 & 75 & 1.3084 & -24.966\\
\midrule
15&58  & 0.28644 & 0.19884  & 4.4155 & 25 & 26.274 & -23.657\\
\midrule
16&59  & 0.29797 & 0.24051  & 18.419 & 50 & 26.274 & -23.657\\
\midrule
17&60  & 0.30272 & 0.2622  & 48.278  & 75 & 26.274 & -23.657\\
\midrule
18&62  & 0.36941 & 0.18108  & 4.4155 & 25  & 51.24 & -22.349\\
\midrule
19&63  & 0.35288 & 0.2259 & 18.419 & 50 & 51.24 & -22.349\\
\midrule
20&64  & 0.40795 & 0.21059 & 18.419 & 50 & 76.206 & -21.041\\
\midrule
21&44  & 0.23784 & 0.35662 & 72.065 & 88 & -49.931 & -2.6168\\
\midrule
22&43 & 0.25332 & 0.35108 & 18.419 & 50 & -24.966 & -1.3084\\
\midrule
23&42 & 0.26938 & 0.34523 & 48.278 & 75 & -24.966 & -1.3084\\
\midrule
24&41 & 0.27473 & 0.34327 & 72.065 & 88 & -24.966 & -1.3084\\
\midrule
25&6 & 0.313 & 0.3290 & 0 & 0 & 0 & 0\\
\midrule
26&5 & 0.31273 & 0.32902 & 4.4155 & 25 & 0 & 0\\
\midrule
27&4 & 0.31273 & 0.32902 & 18.419 & 50 & 0 & 0 \\
\midrule
28&3 & 0.31273 & 0.32902 & 48.278 & 75 & 0 & 0 \\
\midrule
29&1 & 0.31273 & 0.32902 & 100.00 &  100 & 0 & 0 \\
\midrule
30&2 & 0.31273 & 0.32902 & 72.065 & 88 & 0 & 0 \\
\midrule
31&67 & 0.41044 & 0.2905 & 4.4155 & 25 & 24.966 & 1.3084 \\
\midrule
32&68 & 0.37353 & 0.30534 & 18.419 & 50 & 24.966 & 1.3084\\
\midrule
33&69 & 0.3568 & 0.31196 & 48.278 & 75 & 24.966 & 1.3084 \\
\midrule
34&66 & 0.43376 & 0.28095 & 18.419 & 50 & 49.931 & 2.6168 \\
\midrule
35&65 & 0.49181 & 0.25666  & 18.419 & 50 & 74.897 & 3.9252\\
\midrule
36 &40& 0.27022 & 0.43268 & 48.278 & 75 & -51.24 & 22.349 \\
\midrule
37&39 & 0.27623 & 0.41829 & 72.065 & 88 & -51.24 & 22.349 \\
\midrule
38 &36& 0.3075 & 0.52435 & 4.4155 & 25  & -26.274 & 23.657\\
\midrule
39 &33& 0.31561 & 0.44408 & 18.419 & 50  &  -26.274 & 23.657\\
\midrule
40 &32& 0.31654 & 0.41004 & 48.278 & 75 & -26.274 & 23.657\\
\midrule
41&31 & 0.31652 & 0.39917 & 72.065 & 88 & -26.274 & 23.657\\
\midrule
42&20 & 0.41791 & 0.4496 & 4.4155 & 25 & -1.3084 & 24.966\\
\midrule
43&17 & 0.38179 & 0.40728 & 18.419 & 50 & -1.3084 & 24.966\\
\midrule
44&16 & 0.36355 & 0.38634 & 48.278 & 75 & -1.3084 & 24.966\\
\midrule
45&15 & 0.35735 & 0.37928 & 72.065 & 88 & -1.3084 & 24.966\\
\midrule
46 &10& 0.52174 & 0.37656 & 4.4155 & 25 & 23.657 & 26.274\\
\midrule
47 &9& 0.44682 & 0.36993 & 18.419 & 50 & 23.657 & 26.274\\
\midrule
48 &8& 0.41032 & 0.36214 & 48.278 & 75 & 23.657 & 26.274\\
\midrule
49 &7& 0.50873 & 0.33341 & 18.419 & 50 & 48.623 & 27.583\\
\midrule
50 &70 & 0.56618 & 0.29873 & 18.419 & 50 & 73.589 & 28.891\\
\midrule
51&38 & 0.29736 & 0.57731 & 18.419 & 50 & -52.548 & 47.315\\
\midrule
52&35 & 0.31049 & 0.50294 & 48.278 & 75 & -52.548 & 47.315\\
\midrule
53 &34& 0.31313 & 0.47888 & 72.065 & 88 & -52.548 & 47.315\\
\midrule
54 &28& 0.36753 & 0.52508 & 18.419 & 50 & -27.583 & 48.623\\
\midrule
55 &28& 0.35998 & 0.47135 & 48.278 & 75 & -27.583 & 48.623\\
\midrule
56&26 & 0.35593 & 0.45307 & 72.065 & 88 & -27.583 & 48.623\\
\midrule
57 & 22& 0.43671 & 0.47238 & 18.419 & 50 & -2.6168 & 49.931\\
\midrule
58 &19& 0.4092 & 0.43925 & 48.278 & 75 & -2.6168 & 49.931\\
\midrule
59 &18& 0.39861 & 0.42682 & 72.065 & 88 & -2.6168 & 49.931\\
\midrule
60&13 & 0.50246 & 0.42134 & 18.419 & 50 & 22.349 & 51.24\\
\midrule
61&12 & 0.4572 & 0.40735 & 48.278 & 75 & 22.349 & 51.24\\
\midrule
62 &11 & 0.56315 & 0.37345 & 18.419 & 50 & 47.315 & 52.548\\
\midrule
63 &71& 0.61793 & 0.32963 & 18.419 & 50 & 72.28 & 53.857\\
\midrule
64 &37& 0.30023 & 0.56426 & 72.065 & 88 & -78.822 & 70.972\\
\midrule
65 &29& 0.34264 & 0.56147 & 48.278 & 75 & -53.857 & 72.28\\
\midrule
66&30 & 0.34455 & 0.53221 & 72.065 & 88 & -53.857 & 72.28\\
\midrule
67 &24& 0.39381 & 0.52123 & 48.278 & 75 & -28.891 & 73.589\\
\midrule
68&25 & 0.38886 & 0.49964 & 72.065 & 88 & -28.891 & 73.589\\
\midrule
69 &23& 0.44388 & 0.48131 & 48.278 & 75 & -3.9252 & 74.897\\
\midrule
70&21 & 0.43244 & 0.46714 & 72.065 & 88 & -3.9252 & 74.897\\
\midrule
71&14 & 0.49196 & 0.4425 & 48.278 & 75 & 21.041 & 76.206\\
\bottomrule
\caption{  Coordinates for the University of Wisconsin 71 (UW-71) colors in CIE 1931 xyY space and CIELAB color space. The white point used to convert between CIE 1931 xyY and CIELAB space was CIE Illuminant D65 (x = 0.313, y = 0.329, Y = 100). The 'Sorted Position' column indicates the index of the color when the colors are sorted by hue angle and chroma.}
\label{tab:UW_71_colors}
\end{longtable}
}

\end{document}